\documentclass[preprint,12pt]{elsarticle}

\usepackage{mathrsfs}
\usepackage{blindtext}
\usepackage{algorithm2e}      
\usepackage{lineno}
\usepackage{graphicx}
\usepackage[utf8]{inputenc}
\usepackage{lmodern}
\usepackage[T1]{fontenc}
\usepackage{lscape}
\usepackage{amsmath}
\usepackage{amssymb}
\usepackage{cases}
\usepackage{graphicx}
\usepackage{caption}
\usepackage{comment}

\usepackage{makecell}
\usepackage{multirow}
\usepackage{amsfonts}
\usepackage{float} 
\usepackage{diagbox}
\usepackage{fullpage}
\usepackage{url}
\usepackage{rotating}
\usepackage{eurosym}
\usepackage{wrapfig}
\usepackage[final]{pdfpages}
\usepackage{epstopdf} 
\usepackage[a4paper]{geometry}
\usepackage[nottoc]{tocbibind}
\usepackage{placeins}
\usepackage{float}
\usepackage{listings}
\usepackage{color, colortbl}
\usepackage{hyperref}
\usepackage{xcolor}
\usepackage{graphicx, caption, subcaption}
\usepackage{acronym}

\usepackage{sidecap}
\hypersetup{
colorlinks,
citecolor=black,
filecolor=black,
linkcolor=black,
urlcolor=black
}

\newcommand{\hao}{\textcolor{black}}

\journal{Computer Methods in Applied Mechanics and Engineering}
\graphicspath{{./figures/}}

\begin{document}
\newgeometry{top=4cm}
\begin{frontmatter}

\title{Improving Long-term Autoregressive Spatiotemporal Predictions: A Proof of Concept with Fluid Dynamics}

\author[inst1]{Hao Zhou}

\affiliation[inst1]{organization={School of Mechanical, Medical and Process Engineering, Faculty of Engineering, Queensland University of Technology },
            country={AU}}

\author[inst2]{Sibo Cheng\corref{cor1}}



\cortext[cor1]{corresponding: sibo.cheng@enpc.fr}

\affiliation[inst2]{organization={CEREA, ENPC, EDF R\&D, Institut Polytechnique de Paris, \^Ile-de-France },
country={France}}


\begin{abstract}
Data-driven approaches have emerged as a powerful alternative to traditional numerical methods for forecasting physical systems, offering fast inference and reduced computational costs. However, for complex systems and those without prior knowledge, the accuracy of long-term predictions frequently deteriorates due to error accumulation. Existing solutions often adopt an autoregressive approach that unrolls multiple time steps during each training iteration; although effective for long-term forecasting, this method requires storing entire unrolling sequences in GPU memory, leading to high resource demands. Moreover, optimizing for long-term accuracy in autoregressive frameworks can compromise short-term performance. To address these challenges, we introduce the Stochastic PushForward (SPF) training framework in this paper. SPF preserves the one-step-ahead training paradigm while still enabling multi-step-ahead learning. It dynamically constructs a supplementary dataset from the model’s predictions and uses this dataset in combination with the original training data. By drawing inputs from both the ground truth and model-generated predictions through a stochastic acquisition strategy, SPF naturally balances short- and long-term predictive performance and further reduces overfitting and improves generalization. Furthermore, the training process is executed in a one-step-ahead manner, with multi-step-ahead predictions precomputed between epochs—thus eliminating the need to retain entire unrolling sequences in memory, thus keeping memory usage stable. \hao{We demonstrate the effectiveness of SPF on the Burgers' equation and the Shallow Water benchmark.} Experimental results demonstrated that SPF delivers superior long-term accuracy compared to autoregressive approaches while reducing memory consumption. This positions SPF as a promising solution for resource-constrained environments and complex physical simulations.
\end{abstract}


\end{frontmatter}
\newgeometry{top=2.5cm}
\section*{Main Notations} 
\begin{tabular}{ll}
\hline
\textbf{Notation} & \textbf{Description} \\
\hline
\\
  & \textit{Baseline Method} \\
  \\
$\mathit{\mathbf{x}}_{t}$ & State vector in the full space at time t \\
$\boldsymbol{\eta}_{t}$ & Compressed state vector in the latent space at time t\\
$\mathit{\mathbf{x}}^r_{t}$ & Reconstruction state vector in the full space at time t \\
$\mathcal{F}_e, \mathcal{F}_d$ & Encoder, Decoder function in autoencoder\\
$\theta_{\mathcal{F}_e}, \theta_{\mathcal{F}_d}$ & Parameters for the encoder and decoder \\
$T$ & Total number of time steps in dataset \\
$f$ & Surrogate model for latent dynamics \\
$\tilde{\boldsymbol{\eta}}_t$ & Output of surrogate model in the latent space at time t\\
$\mathcal{F}_{\textrm{LSTM}}$ & LSTM function \\
$k_\textrm{in}, k_\textrm{out}$ & Input and Output time steps of LSTM \\
${\boldsymbol{\eta}}_{t:t+k_\textrm{in}-1}$ & Sequence of compressed state vectors \\
$\mathcal{L}$ & Loss function \\
$\delta$ & Depth of the autoregressive progress \\
$f^\delta$ & Function composed with itself $\delta$ times \\
$\lambda_\delta$ & Coefficient to control the influence of $\delta$-step predictions\\
$f^*$ & Frozen predictive model \\
\\
  & \textit{Stochastic PushForward} \\
  \\
$\mathcal{D}_1$ & Original Dataset\\
$\mathcal{D}_\delta$ & Supplementary Dataset\\
$\mathcal{D}$ & Integration of original and supplementary Dataset\\
$I$ & Random variable follows Bernoulli distribution \\
$p$ & Probability of drawing data from $\mathcal{D}_1$\\
$\eta^I$ & Input data sampled by acquisition method \\
$\eta^{\mathcal{D}}_{t}$ & Data at time t drawing from $\mathcal{D}$\\
$\gamma$ & Weight coefficient for data from different dataset\\
$\alpha$ & Weight coefficient for data from supplementary dataset\\
$N_\text{init}$ & Max. initial predictive model training epoch \\
$N_\text{epoch}$ & Max. sequential predictive model training epoch \\
$N_\text{UI}$ & Supplementary dataset update interval\\

\\
  & \textit{\hao{Test cases}} \\
  \\
$V$ & Flow speed for Burgers' equation\\
$h$ & Total water depth including the undisturbed water depth\\
$u, v$ & Velocity components in the x (horizontal) and y (vertical) directions\\
$g$ & Gravitational acceleration\\
$r$ & Spatial euclidean distance\\
$\epsilon$ & Balgovind type of correlation function \\
$L$ & Typical correlation length scale\\
\hline
\end{tabular}
\restoregeometry
\section{Introduction}

Over many years, scientific research has produced highly detailed mathematical models of physical phenomena\cite{tabatabaei_techniques_2022}. These models are frequently and naturally expressed in the form of differential equations~\cite{olver2014introduction}, most commonly as time-dependent \ac{PDE}s. Solving these equations plays a pivotal role in a wide array of fields, including meteorology, oceanography, and engineering. Traditional methods, including \ac{FDM}~\cite{casulli_semi-implicit_1990,kurganov_central-upwind_2002}, \ac{FVM}~\cite{alcrudo_high-resolution_1993, bale2003wave} and \ac{LBM}~\cite{qian_lattice_1992, shan_lattice_1993}, have been proven reliable for achieving high-fidelity and high-accuracy results. However, the slow computational speed and the requirement of significant resources make it less ideal in high-dimensional systems, especially in real-time predictions~\cite{ babanezhad_functional_2020, lagha_body_2023, zuo2009real}.

Researchers have turned to employ \ac{ML} and \ac{DL} techniques~\cite{fresca2021real, drakoulas2023fastsvd} to predict the system evolution due to their quick prediction capabilities and low cost after training. A wide range of neural-network-based sequence models can serve in this surrogate role—examples include recurrent networks~\cite{elman1990finding}, transformers~\cite{yu2023dsformer,young2022dateformer}, and other modern architectures~\cite{lara2021experimental}. These models are typically trained using a one-step-ahead prediction approach, where the model learns to predict the next state in a sequence based on the past state, with iterative prediction used to achieve longer forecasting horizons.

However, despite their strengths, these surrogate models often encounter challenges in long-term predictions due to the accumulation of errors over time. To mitigate this issue, researchers have proposed various methods to enhance the long-term prediction abilities~\cite{chen2024multi,trehan2017error,cheng2024multi}. One approach involves the use of encoder-decoder architectures, which effectively capture complex temporal patterns over the sequences~\cite{fukami2021machine, hasegawa2020machine}. For example, Maulik et al.~\cite{maulik2021reduced} developed a \ac{CAE}-\ac{LSTM} model demonstrating that incorporating a dimensionality reduction step leads to more stable and accurate long-term predictions. Another strategy is to integrate physical constraints into the learning process, enabling the model to adhere to known physical laws and reduce error propagation~\cite{raissi2019physics, karniadakis2021physics, wang2024dynamical, xiao2025meshless}. Our previous work~\cite{zhou2024multi} leveraged physical constraints in a multi-fidelity framework, employing low-fidelity fields to guide the training of high-fidelity models, and achieved a 50\% improvement in \ac{MSE} compared to the original model.
Nevertheless, employing these methods has their limitations. For instance, physical constraints may not be suitable in cases where there is a lack of prior knowledge about the control laws~\cite{rudy2017data,raissi2018deep} or in large-scale scenarios~\cite{de2019deep,chattopadhyay2020data} where integrating detailed physical laws becomes computationally infeasible. 

To overcome these problems, researchers have introduced multi-step-ahead prediction methods, notably within the \ac{ATF}~\cite{bielitz2023identification,vlachas2023learning}. In this approach, the model's output is recursively fed back as inputs for subsequent time steps. During training, losses are integrated over these multiple prediction steps, compelling the model to learn representations that are robust over extended sequences and capable of capturing long-term dependencies~\cite{wu2023learning}. 
However, the \ac{ATF} comes with significant drawbacks. One of the primary challenges is the high demand for GPU memory. The recrusive nature of feeding outputs back into the model requires storing intermediate states and gradients for each time step during training. This accumulation can quickly exhaust available memory, especially with long sequences or high-dimensional data. Additionally, the increased weights on long-term predictions often lead to diminished short-term prediction accuracy~\cite{chattopadhyay2020data}. To address these challenges, Brandstetter et al.~\cite{brandstetter2022message} proposed a \ac{PF} method as an enhancement to \ac{ATF}. It introduced a stability loss term to mitigate distribution shifts during long rollouts. It unrolls the model over multiple steps but backpropagates only through the final step, reducing memory usage while improving robustness against error accumulation. Additionally, their implementation allows the rollout length to be dynamically adjusted during training. 

\begin{figure}
    \hspace*{-0.5cm}
    \centering
    \includegraphics[width=0.9\linewidth]{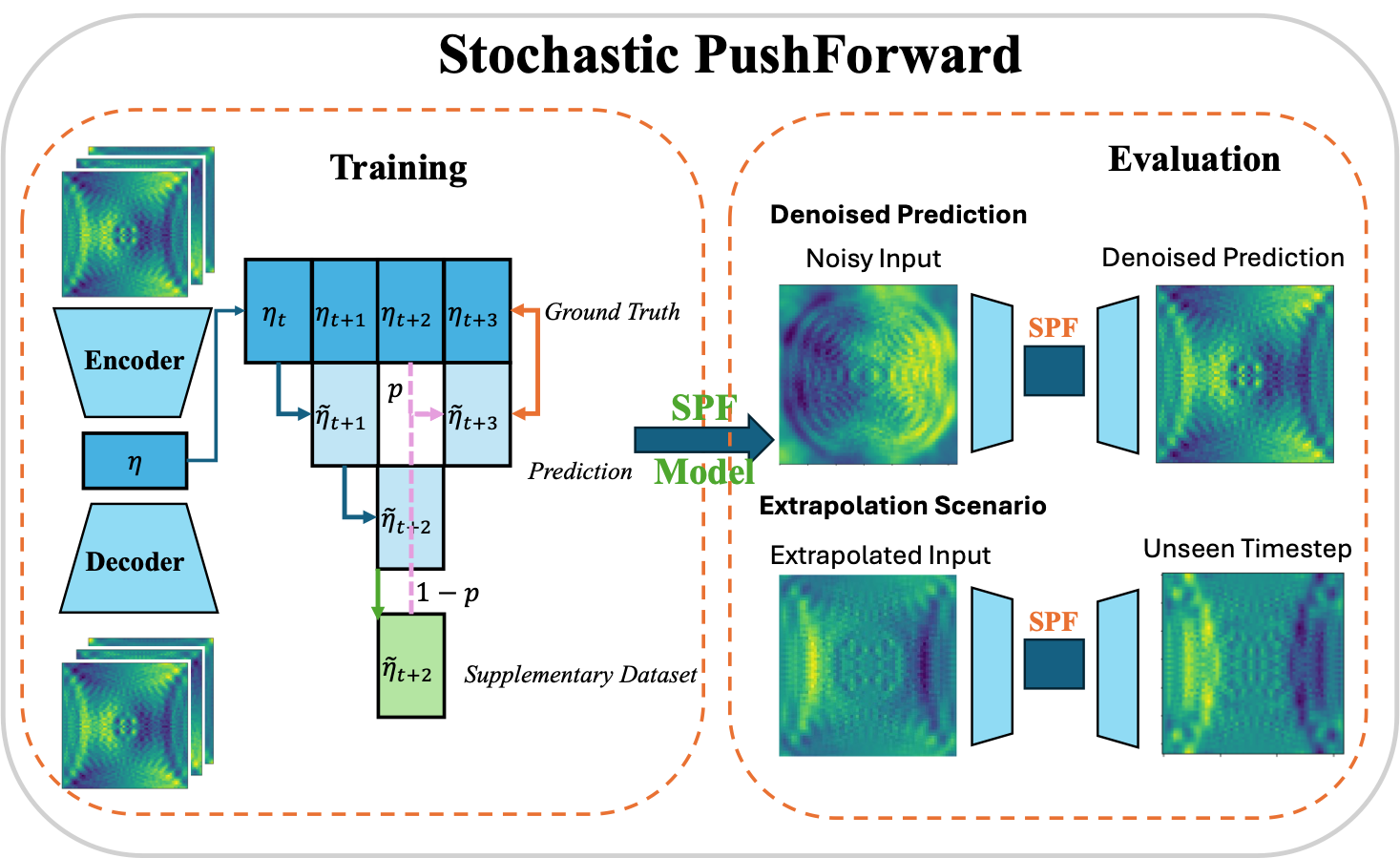}
    \caption{Illustration of training and evaluation process for \ac{SPF}}
    \label{fig:training_evaluation}
\end{figure}

Inspired by the \ac{PF} method, we propose a \ac{SPF} method. The key distinction in training process between \ac{SPF} and \ac{PF} is that in \ac{SPF}, the final backpropagation step can use inputs drawn either from prediction results or from ground truth, whereas \ac{PF} relies solely on its own predictions. Additionally, \ac{SPF} introduces a supplementary dataset that keeps the one-step-ahead training strategy, which can further reduce memory usage compared to \ac{PF}. The supplementary dataset in \ac{SPF} is generated dynamically during training, storing the model’s predictions for use as inputs in subsequent training iterations. Unlike traditional autoregressive training—where the model must perform multiple unrolling steps in every iteration—\ac{SPF} carries out it between epochs.  This design preserves training simplicity while still capturing the benefits of multi-step-ahead predictions. After a predefined number of epochs, the model generates new predictions for the initial dataset, thereby updating the supplementary dataset. Then \ac{SPF} employs the stochastic acquisition method to select data from both the initial dataset and the supplementary dataset during training. In addition, it limits the size of the dataset used in each training iteration, ensuring stable GPU memory usage. By storing and selectively sampling predictions in the supplementary dataset, \ac{SPF} eliminates the need to retain full unrolling sequences in memory during training, in contrast to \ac{PF}, which must backpropagate through multiple time steps. This reduction in memory requirements makes \ac{SPF} particularly suitable for long-term prediction tasks or resource-constrained environments. To validate its effectiveness, we apply \ac{SPF} to the Burgers' equation and Shallow water benchmark~\cite{cheng2025machine}. The results underscore \ac{SPF}’s capacity to improve predictive accuracy under complex and dynamic conditions regarding several evaluation metrics. The overall training and evaluation processes integrating with \ac{CAE} are shown in the Fig.~\ref{fig:training_evaluation}. \hao{The developed model is evaluated on test data generated from initial conditions and physical parameters that differ from those used during training.} \hao{Furthermore, the evaluation includes unseen time steps beyond the training horizon, allowing the model’s performance to be assessed under extrapolative scenarios.} 

In summary, we make the following main contributions in this study:
\begin{enumerate}
    \item We propose a novel training framework for improving long-term prediction, named \ac{SPF}. By leveraging a supplementary dataset, \ac{SPF} achieves multi-step-ahead training via a one-step-ahead approach, thereby enhancing long-term prediction accuracy and reducing computational resource demands, thus improving model frugality.
    \item By integrating and periodically updating the supplementary dataset, \ac{SPF} effectively mitigates the impact of limited training data. This dynamic data augmentation approach enables the model to maintain and even improve long-term prediction accuracy, despite scarce or incomplete datasets.
    \item \hao{When evaluated on two benchmark cases, the proposed \ac{SPF} consistently outperforms the baseline LSTM, as well as the \ac{ATF} and \ac{PF} methods.} The model achieves superior results on \ac{MSE} and \ac{SSIM} metrics for long-term predictions and noisy data, confirming its ability to deliver more accurate, perceptually consistent and robust forecasts over extended time horizons.
\end{enumerate}

The rest of this paper is organised as follows. In Section~\ref{sec:BaselineMethods}, we introduce the state-of-the-art \ac{ATF} and \ac{PF} for long-term prediction and essential components. Section~\ref{sec:SPF} presents the structure of \ac{SPF} and details the training methodology for it. The numerical experiments, Burger' system and Shallow water system, are discussed in Section~\ref{sec:burger} and Section~\ref{sec:sw} separately. Finally, we conclude and summarise our findings in Section~\ref{sec:conclusion}.

\section{Baseline Methods: Autoregressive Framework and PushForward for Long-term Prediction}
\label{sec:BaselineMethods}


In this section, we will introduce two baseline training methods: an autoregressive training framework and the pushforward training paragdigm. The autoregressive approach combines losses from multiple unrolled steps to capture long-term dependencies. Building on this, \ac{PF} employs a stability loss to mitigate distribution shifts, and unrolls multiple steps while only backpropagating through the final step. 

Before diving into these specifics of the methods, it is important to first introduce the key components that form the foundation of this approach: \ac{AE} and surrogate models. \ac{AE}s are used to compress high-dimensional data into a compact representation, reducing complexity and noise.

\subsection{Reduced Order Modelling: Autoencoder}

An \ac{AE} is a type of neural network designed to reduce the dimensionality of high-dimensional data while retaining its essential features~\cite{hinton2006reducing}. It operates through an encoder-decoder framework. The encoder, \( \mathcal{F}_e \), compresses the input \( \mathbf{x}_t \in \mathbb{R}^n \) at a given time step $t$ into a lower-dimensional latent vector \( \boldsymbol{\eta}_t \in \mathbb{R}^m \), and the decoder, \( \mathcal{F}_d \), reconstructs the original state \( \mathbf{x}^r_t \in \mathbb{R}^n \) from this latent space, where $n, m$ are the dimensions of the full and the reduced state vector, respectively. During training, the encoder and decoder are jointly optimized in an end-to-end manner to minimize the reconstruction error. Using the \ac{MSE} as the loss function \( \mathcal{L}(\cdot) \), the process can be described as:

\begin{equation}
    \label{eq:CAE_combined}
    \boldsymbol{\eta}_{t} = \mathcal{F}_e(\mathbf{x}_{t}), \quad \mathbf{x}^r_{t} = \mathcal{F}_d(\boldsymbol{\eta}_{t}), \quad
    \mathcal{L}(\theta_{\mathcal{F}_e}, \theta_{\mathcal{F}_d}) = \frac{1}{T} \sum_{i=1}^{T} \bigl\lVert \mathbf{x}^r_i - \mathbf{x}_i \bigl\rVert_2^2
\end{equation}

In this setup, \( \theta_{\mathcal{F}_e} \) and \( \theta_{\mathcal{F}_d} \) represent the parameters of the encoder and decoder, respectively, while \( T \) denotes the total number of time steps, and \( \parallel \cdot \parallel_2 \) refers to the Euclidean norm.

\subsection{Surrogate Model for Latent Dynamics}
\label{subsec:surrogate_model_alt}

After obtaining the latent representation $\boldsymbol{\eta}_{t}$ from the original data $\mathbf{x}_{t}$ (e.g., via an \ac{AE}), the next step is to capture the system’s temporal evolution for time series prediction. Within a \ac{ROM} framework, this typically means modeling how the latent representation evolves over time. Any neural-network-based sequence model—such as recurrent networks (\ac{RNN}~\cite{rumelhart1986learning} or \ac{LSTM}~\cite{graves2012long}), Transformers~\cite{vaswani2017attention}, or other architectures—can serve as this surrogate model.

\begin{figure}
    \hspace*{-1.5cm}
    \centering
    \includegraphics[width=1.2\linewidth]{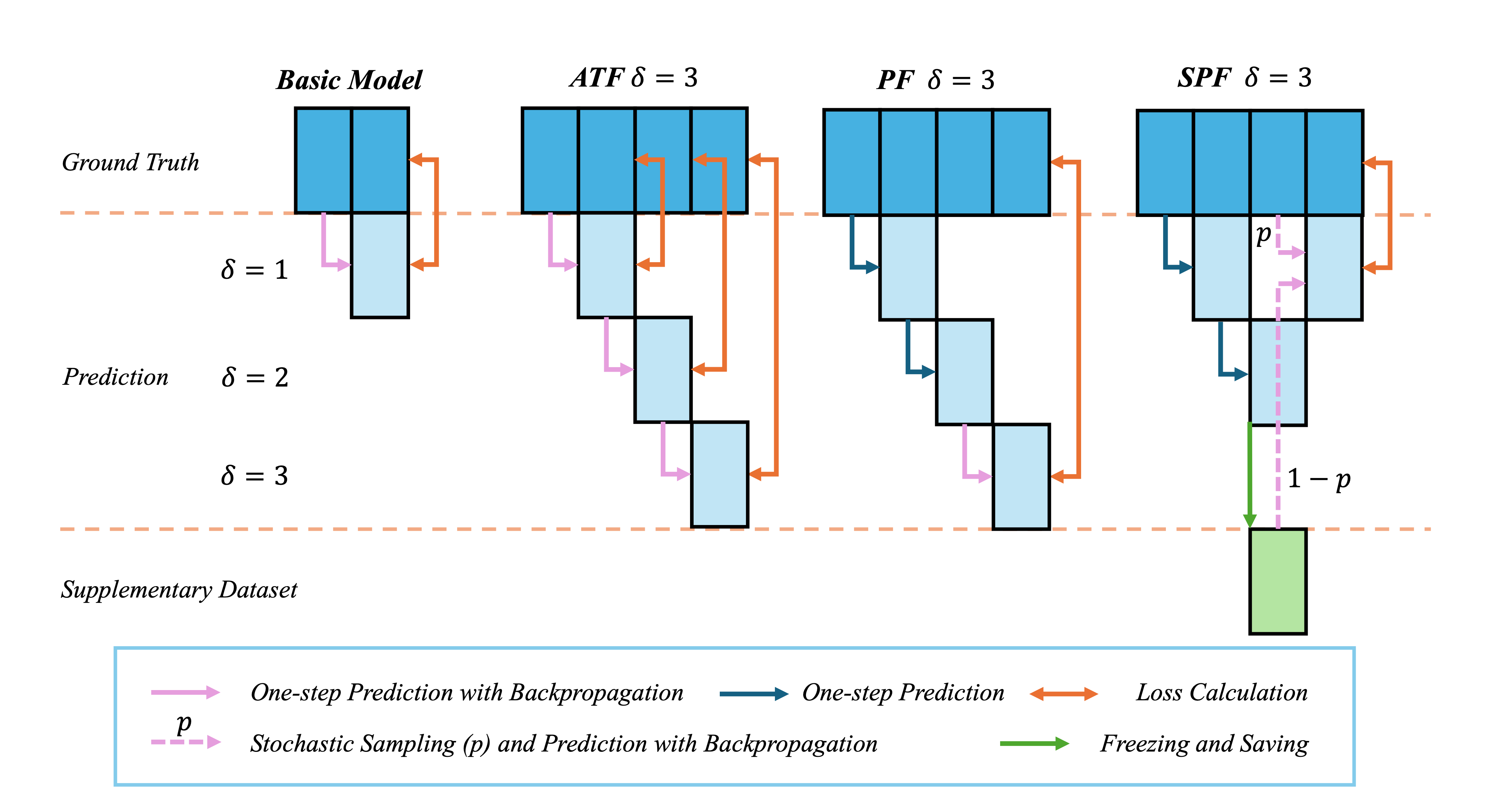}
    \caption{Illustration and comparison of different training framework}
    \label{fig:frameworks_compar}
\end{figure}

For time series that encode latent representations \( [{\boldsymbol{\eta}}_{1}, {\boldsymbol{\eta}}_{2}, \dots, {\boldsymbol{\eta}}_T] \), the surrogate model is trained by progressively shifting the starting time step. The training process can be expressed as:

\begin{equation}
\label{eq:onestep_predict_alt}
\widetilde{\boldsymbol{\eta}}_{t+1} = f\bigl(\boldsymbol{\eta}_{t}\bigr).
\end{equation}

where $f(\cdot)$ represents the surrogate model and $\tilde{\boldsymbol{\eta}}_{t+1}$ is the predicted latent state at time $t+1$. During training, the difference between the predicted and true latent states is measured by a loss function (e.g., \ac{MSE}):

\begin{equation}
\label{eq:onestep_loss}
\mathcal{L}_t(\theta_f) \;=\; \bigl\lVert \tilde{\boldsymbol{\eta}}_{t+1} - \boldsymbol{\eta}_{t+1} \bigr\rVert_2^2 = \bigl\lVert f(\boldsymbol{\eta}_t) - \boldsymbol{\eta}_{t+1} \bigr\rVert_2^2 ,
\end{equation}


and the model parameters are updated to minimize this loss. The total process is performed in Fig.~\ref{fig:frameworks_compar}.

Although one-step training is straightforward, some tasks benefit from a \emph{multi-step} or \ac{Seq2Seq} setup, which takes $k_{\textrm{in}}$ consecutive latent states and outputs the next $k_{\textrm{out}}$ states:
\begin{equation}
\label{eq:seq2seq_predict_alt}
\widetilde{\boldsymbol{\eta}}_{t + k_{\textrm{in}} :\, t + k_{\textrm{in}} + k_{\textrm{out}} - 1}
= \mathcal{F}_\textrm{LSTM}\bigl(\boldsymbol{\eta}_{t :\, t + k_{\textrm{in}} - 1}\bigr).
\end{equation}
where $\mathcal{F}_{\textrm{LSTM}}$ is a predictive model (an \ac{LSTM} in this example), and $k_{\textrm{in}}, k_{\textrm{out}}$ denote the lengths of the input and output sequences, respectively. The loss function in this case measures the discrepancy between $\widetilde{\boldsymbol{\eta}}_{t + k_{\textrm{in}} :\, t + k_{\textrm{in}} + k_{\textrm{out}} - 1}$ and its ground-truth counterpart over all training samples. In our numerical experiments, we employ an \ac{LSTM}-based \ac{Seq2Seq} setup, but to keep the framework more general and more easily explained, the notation of one-step training is primarily used in this methodology section.

Once trained, the surrogate model can perform extended forecasts through an iterative (or circular) procedure. Specifically, after predicting $\widetilde{\boldsymbol{\eta}}_{t+1}$, the model feeds that newly estimated state back into its input to predict $\widetilde{\boldsymbol{\eta}}_{t+2}$, and so on:
\begin{equation}
\label{eq:LSTM_Predict_alt}
\widetilde{\boldsymbol{\eta}}_{t+1} =
\begin{cases} 
f\bigl(\boldsymbol{\eta}_{t}\bigr), & n=0, \\
f\bigl(\widetilde{\boldsymbol{\eta}}_{t}\bigr), & n>0,
\end{cases}
\quad 
t = n + 1,\quad n = 0,1,2,\dots
\end{equation}

At $n=0$, the model uses the ground-truth state $\boldsymbol{\eta}_{t}$; for subsequent steps, it uses its own previous prediction $\widetilde{\boldsymbol{\eta}}_{t}$. By using this iterative prediction approach, the surrogate model can provide long-term forecasts.

\subsection{Autoregressive Training Framework}

In the traditional time-series training process, as shown in Eq.~\eqref{eq:onestep_predict_alt} and ~\eqref{eq:onestep_loss}, only one prediction step is considered at a time. This approach tends to focus primarily on short-term, single-step-ahead predictions, causing the surrogate model to overfit to immediate transitions between consecutive steps while neglecting some underlying patterns required for accurate long-term predictions. Therefore, to enhance the model's long-term prediction accuracy, it is intuitive to include the loss from multiple prediction steps rather than just one in the final loss function. This naturally leads to the adoption of a modified autoregressive training framework~\cite{engle1982autoregressive}. For example, as shown in Eq.~\eqref{eq:Rollout_Train}, starting with an initial input sequence ${\boldsymbol{\eta}}_{t}$, the model predicts future states over two main processes: a one-step-ahead prediction and a $\delta$-step-ahead prediction. The corresponding errors are calculated for these two processes.


\begin{equation}
\label{eq:Rollout_Train}
\begin{cases} 
\tilde{\boldsymbol{\eta}}_{t+1} = f(\boldsymbol{\eta}_{t}), & \text{One-step-ahead prediction} \\
\tilde{\boldsymbol{\eta}}_{t+2} = f(f(\boldsymbol{\eta}_{t})), & \text{Two-step-ahead prediction} \\
\tilde{\boldsymbol{\eta}}_{t+\delta} = f^\delta(\boldsymbol{\eta}_{t}), & \delta\text{-step-ahead prediction} \\
f^\delta(\boldsymbol{\eta}_t) = 
\begin{cases} 
\boldsymbol{\eta}_t, & \text{if } \delta = 0, \\
f(f^{\delta-1}(\boldsymbol{\eta}_t)), & \text{if } \delta > 0,
\end{cases} & \text{Definition of iterative composition.}
\end{cases}
\end{equation}

where $f^\delta$ represents the function $f$ composed with itself $\delta$ times, as shown in the definition of recursive composition.

The loss function based on the autoregressive training process can now be expressed as:

\begin{equation}
\label{eq:Rollout_Loss_Updated}
\begin{aligned}
    \mathcal{L}_t(\theta_f) &= \parallel \tilde{\boldsymbol{\eta}}_{t+1} - {\boldsymbol{\eta}}_{t+1} \parallel_2^2 + \parallel \tilde{\boldsymbol{\eta}}_{t+2} - {\boldsymbol{\eta}}_{t+2} \parallel_2^2 + \cdots + \parallel \tilde{\boldsymbol{\eta}}_{t+\delta} - {\boldsymbol{\eta}}_{t+\delta} \parallel_2^2 \\
    &= \parallel f({\boldsymbol{\eta}}_{t}) - {\boldsymbol{\eta}}_{t+1} \parallel_2^2 + \parallel f^2({\boldsymbol{\eta}}_{t}) - {\boldsymbol{\eta}}_{t+2} \parallel_2^2 + \cdots + \parallel f^\delta({\boldsymbol{\eta}}_{t}) - {\boldsymbol{\eta}}_{t+\delta} \parallel_2^2 \end{aligned}
\end{equation}



where \( \parallel \cdot \parallel_2 \) is the Euclidean norm. For briefly, we only consider a single input $\mathbf{\eta}_{t}$ in Eq. \eqref{eq:Rollout_Loss_Updated}. Fig.~\ref{fig:frameworks_compar} illustrates the structure of \ac{ATF} when $\delta=3$.

In practice, simply summing the loss from both steps might not yield optimal results. Since the time-series model is fundamentally based on single-step-ahead predictions, the loss from the multi-step-ahead predictions (i.e., the $\delta$-step) can be treated as a penalty, rather than being given equal weight to the single-step-ahead loss. To account for this, a coefficient $\lambda_\delta$ for the $\delta$-step-ahead prediction is introduced, where $\lambda_1$ is set to 1 for the single-step-ahead prediction, and $\lambda_\delta$ is set to a value to control the influence of multi-step-head predictions on the training process.

\begin{equation}
\label{eq:Rollout_Loss_Updated_Penalty}
\begin{aligned}
    \mathcal{L}_t(\theta_f) = \parallel f({\boldsymbol{\eta}}_{t}) - {\boldsymbol{\eta}}_{t+1} \parallel_2^2 + \lambda_2 \parallel f^2({\boldsymbol{\eta}}_{t}) - {\boldsymbol{\eta}}_{t+2} \parallel_2^2 + \cdots + \lambda_\delta \parallel f^\delta({\boldsymbol{\eta}}_{t}) - {\boldsymbol{\eta}}_{t+\delta} \parallel_2^2
\end{aligned}
\end{equation}

The autoregressive training process is widely used, but it is limited by the high demand on RAM, particularly GPU memory, as it needs to store outputs from $\delta$ steps to compute the final loss~\cite{meng2023towards}. This limitation is especially challenging for large-scale simulations~\cite{pascanu2013difficulty}, as the space complexity increases linearily regarding $\mathcal{O}(\delta)$. However, the memory requirements can often be affordable if the depth $\delta$ is not too large. The real challenge lies in backpropagation, which requires computation of gradients across all these steps. This depth of backpropagation can exacerbate issues like vanishing or exploding gradients, making it difficult for the model to learn long-term dependencies effectively~\cite{pascanu2013difficulty}. Additionally, as gradients propagate over many steps, the training process becomes increasingly unstable. These drawbacks are particularly pronounced in multi-step prediction frameworks, where the accumulation of errors and the need for large memory resources further strain the model's performance.

\subsection{PushForward}


The \ac{PF} method adopts the same training procedure as the autoregressive framework in Eq.~\eqref{eq:Rollout_Train}, but differs in its backpropagation strategy. Specifically, in \ac{PF}, only the final step of the \(\delta\)-step prediction is used to compute gradients for backpropagation. This drastically reduces memory usage compared to backpropagating through all \(\delta\) steps. The loss function is given by:

\begin{equation}
\label{eq:Rollout_Loss_PF}
\begin{aligned}
    \mathcal{L}_t(\theta_f) &=  \parallel \tilde{\boldsymbol{\eta}}_{t+\delta} - {\boldsymbol{\eta}}_{t+\delta} \parallel_2^2 \\
    &= \parallel f(f^{\delta-1, *}({\boldsymbol{\eta}}_{t})) - {\boldsymbol{\eta}}_{t+\delta} \parallel_2^2
\end{aligned}
\end{equation}

where $f^*$ denotes a frozen predictive model, meaning its parameters are fixed, and it is used only for generating data without participating in backpropagation. And the illustration of \ac{PF} when $\delta=3$ is shown in Fig~\ref{fig:frameworks_compar}.

Moreover, \ac{PF} introduces a mechanism for adjusting the rollout length $\delta$ dynamically. A maximum rollout length $\delta_\text{max}$ is set, and at each training step, a random value within the range $[1, \delta_\text{max}]$ is chosen. This variability helps the model learn to generalize more effectively across different rollout scenarios without incurring the same memory demands as a full multi-step-ahead backpropagation. 

While \ac{PF} mitigates many challenges related to memory consumption and gradient instability by only backpropagating through the final step, this approach also makes it highly sensitive to errors from earlier predictions in each training iteration. Because \ac{PF} relies solely on its own outputs, these accumulated errors may bias the learning process and limit improvements for long-term forecasts. Therefore, to overcome these issues, we propose the Stochastic PushForward, which mixes real data with predicted outputs during training to avoid overfitting to noisy predictions and to further enhance long-horizon predictive performance.

\section{Stochastic PushForward for Long-term Prediction}
\label{sec:SPF}

In the previous section, we discussed the Autoregressive Training Framework (ATF) and PushForward (PF) for multi-step-ahead predictions, as well as the use of \ac{AE} for data compression. Building on these foundations, we now introduce \ac{SPF}.

\subsection{Supplementary Dataset}
In SPF, the key idea is to update the original dataset with predicted data. And then the model is trained by simple one-step-ahead training based on the updated dataset as shown in Section~\ref{subsec:surrogate_model_alt}, thereby significantly reducing memory requirements.

Initially, the predictive model is trained using singe-step-ahead prediction on the original dataset $\mathcal{D}_1 = [{\boldsymbol{\eta}}_{1}, {\boldsymbol{\eta}}_{2}, \dots, {\boldsymbol{\eta}}_{T}]$, where ${\boldsymbol{\eta}}_{t}$ represents the data at time $t$, as shown in Eq.~\eqref{eq:initial_LSTM}.

\begin{equation}
\label{eq:initial_LSTM}
    \mathcal{L}_t(\theta_f) = \parallel f(\boldsymbol{\eta}_t) - \boldsymbol{\eta}_{t+1} \parallel_2^2
\end{equation}

Using the trained model, we apply the predictive model $f$ to each ${\boldsymbol{\eta}}_{t}$ in the original dataset to generate the predicted data $\tilde{\boldsymbol{\eta}}_{t+\delta}$:

\begin{equation} 
\tilde{\boldsymbol{\eta}}_{t+\delta} = f^\delta(\boldsymbol{\eta}_t), \quad t = 1, \dots, T-\delta.
\end{equation} 

Using these predictions, we construct a supplementary dataset $\mathcal{D}_{\delta} = [{\boldsymbol{\eta}}_{1}, \dots, {\boldsymbol{\eta}}_{\delta}, \tilde{\boldsymbol{\eta}}_{\delta+1}, \dots, \tilde{\boldsymbol{\eta}}_{T}]$. The overall dataset $\mathcal{D}$ is then formed by combining the original and supplementary datasets:

\begin{equation}
    \mathcal{D} = \mathcal{D}_1  \oplus \mathcal{D}_{\delta}
\end{equation}

By updating the dataset in this manner, we enable the model to learn from both original data and its own predictions with forms: $(\boldsymbol{\eta}_t, \boldsymbol{\eta}_{t+1})$ and $(\tilde{\boldsymbol{\eta}}_t, \boldsymbol{\eta}_{t+1})$, respectively. The integration of supplementary dataset can improve the long-term predictive ability of the model in a similar way with autoregressive framework. Importantly, for each training step, we perform only a simple one-step-ahead training. This reduces the memory demand compared to autoregressive framework.

\subsection{Acquisition Method}

\begin{figure}[ht]
    \centering
    \includegraphics[width=\linewidth]{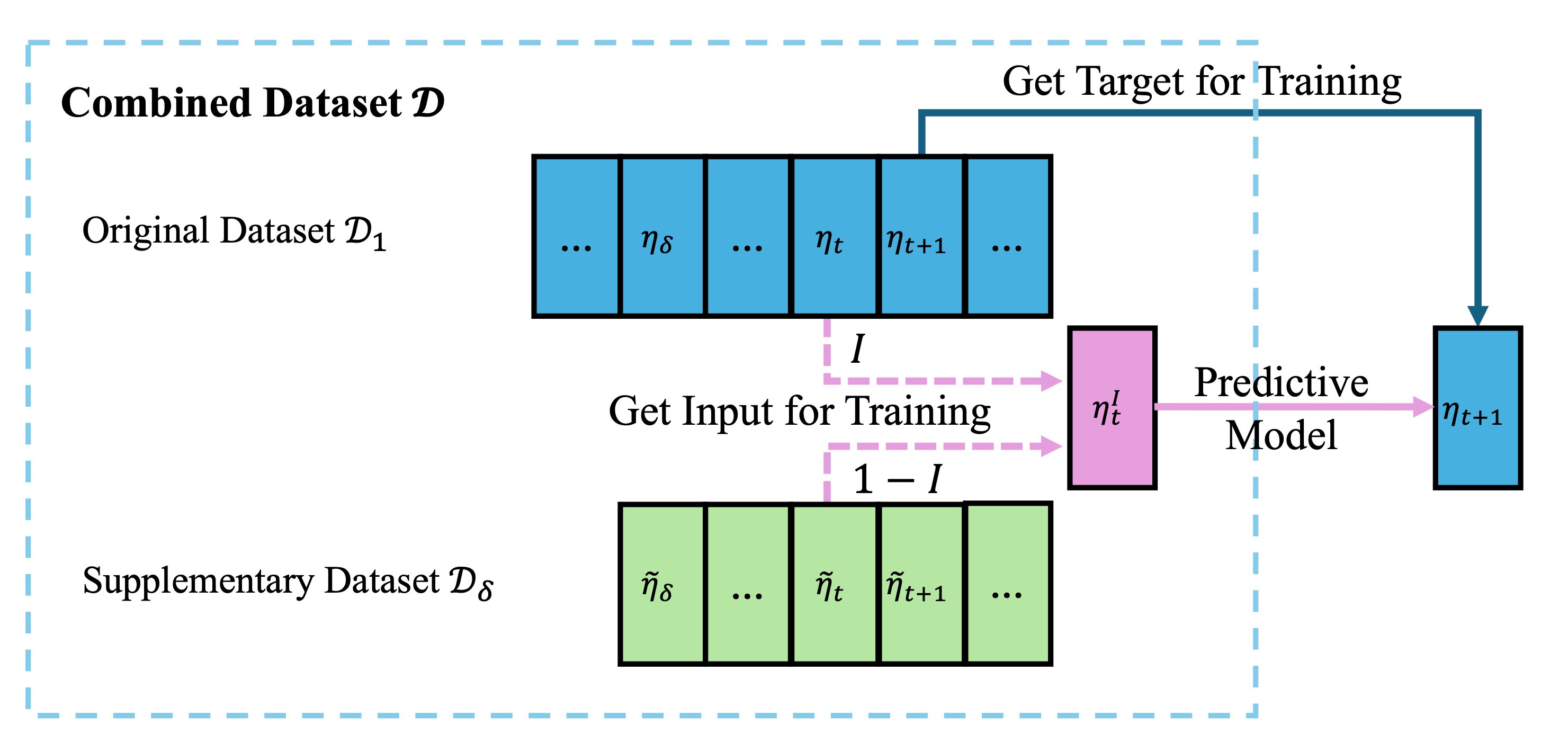} 
    \caption{Illustration of the stochastic acquisition method for training samples from $\mathcal{D}_1$ and $\mathcal{D}_\delta$.}
    \label{fig:acquisition_process}
\end{figure}

To achieve enhancing the model’s long-term predictive capacity while keeping memory usage low and stable, we propose an acquisition method that stochastically selects training samples from both the original dataset $\mathcal{D}_1$ and the supplementary dataset $\mathcal{D}_{\delta}$. This process, as illustrated in Fig.~\ref{fig:acquisition_process}, begins by deciding whether the input $\eta_t$ will come from $\mathcal{D}_1$ or $\mathcal{D}_\delta$ by sampling a random variable $I_t \sim \text{Bernoulli}(p)$.  If $I_t = 1$, then $\eta_t$ is taken from $\mathcal{D}_1$; if $I_t = 0$, it is taken from $\mathcal{D}_\delta$. Regardless of its source, the target $\eta_{t+1}$ is always the corresponding ground truth from $\mathcal{D}_1$.

\begin{equation}
\label{eq:acqfun}
\begin{cases}
I_t \sim \text{Bernoulli}(p), \\
\boldsymbol{\eta}^I_t = I_t \cdot \boldsymbol{\eta}^{\mathcal{D}_1}_{t} + (1 - I_t) \cdot \boldsymbol{\eta}^{\mathcal{D}_{\delta}}_{t}, \\
\boldsymbol{\eta}_{t+1} = \boldsymbol{\eta}^{\mathcal{D}_1}_{t+1}.
\end{cases}
\end{equation}

where superscripts $^{\mathcal{D}_1}$ and $^{\mathcal{D}_{\delta}}$ indicate that $\boldsymbol{\eta}^I_t$ is sampled from $\mathcal{D}_1$ and $\mathcal{D}_{\delta}$, respectively. By mixing real data with model-generated samples, this acquisition method helps maintain good predictive performance for both short- and long-term forecasts.

\subsection{Coefficient in Loss Function}
After sampling a data pair $(\boldsymbol{\eta}^I_t, \boldsymbol{\eta}_{t+1})$, the single-step-ahead training process is applied:
\begin{equation}
    \tilde{\boldsymbol{\eta}}_{t+1} = f(\boldsymbol{\eta}^I_{t})
\end{equation}

The loss for the single-step-ahead training process can be calculated as:
\begin{equation}
\begin{aligned}
\mathcal{L}_t(\theta_f) &= \| \tilde{\boldsymbol{\eta}}_{t+1} - \boldsymbol{\eta}_{t+1} \|^2_2 \\
&= \| f(\boldsymbol{\eta}^I_{t}) - \boldsymbol{\eta}_{t+1} \|^2_2
\end{aligned}
\end{equation}

To balance the influence of samples from the two datasets, we introduce a weighting coefficient $\gamma$ in the loss function. The coefficient adjusts the contribution of the loss from samples where the input comes from $\mathcal{D}_{\delta}$:
\begin{equation}
    \gamma_t =
    \begin{cases}
    1,      & \text{if } \boldsymbol{\eta}^I_t \in \mathcal{D}_1, \\
    \alpha, & \text{if } \boldsymbol{\eta}^I_t \in \mathcal{D}_{\delta}.
    \end{cases}
\end{equation}
where $\alpha$ is a hyperparameter that controls the relative weight of supplementary dataset samples. The total loss over $T$ training steps is:
\begin{equation}
    \mathcal{L}(\theta_f)= \frac{1}{T}  \sum_{t=1}^{T} \gamma_t \mathcal{L}_t
\end{equation}
and the model parameters $\theta_f$ are updated by minimizing $\mathcal{L}(\theta_f)$. This setup allows the training process to incorporate both real and model-generated data while mitigating overfitting to the noisy supplementary samples.
It is important to note that the \ac{SPF} algorithm can be applied with any number of forward depths $\delta$ while in practice it is also possible to train \ac{SPF} progressively with an increasing list of $\delta$. In summary, compared to \ac{ATF} and \ac{PF}, \ac{SPF} reduces the computational burden (particularly the GPU VRAM demand) for multi-step-ahead training by extending the training dataset. More importantly, it mitigates error accumulation in forward predictions by introducing a stochastic sampling technique.   

\subsection{Stochastic PushForward Training Process}

After introducing all the key elements in the \ac{SPF}, the complete training algorithm is demonstrated in Algorithm~\ref{alg:STFinSeq2seq_training}.

\RestyleAlgo{ruled}
\SetKwComment{Comment}{/* }{ */}
\SetEndCharOfAlgoLine{}
\noindent
\begin{minipage}{0.9\linewidth}
\vspace*{-2.0cm}
\hspace*{-1.0cm}
\small
\begin{algorithm}[H]
\caption{Stochastic PushForward Training Process}
\label{alg:STFinSeq2seq_training}

\KwIn{\\
Initial dataset: $\mathcal{D}_1 = [\boldsymbol{\eta}_1, \boldsymbol{\eta}_2, \dots, \boldsymbol{\eta}_{T}]$; \\
Depth: $\delta$; \\
Weighting Coefficient: $\gamma$; \\
Max.~Initial Predictive Model Training Epochs: $N_{\text{init}}$; \\
Max.~Sequential Predictive Model Training Epochs: $N_{\text{epoch}}$; \\
Supplementary Dataset Update Interval: $N_{\text{UI}}$;
}

\BlankLine
\textbf{Procedure: Train Initial Predictive Model}

\For{epoch in $1 \dots N_{\text{init}}$}{
  \For{$t$ in $1 \dots T-1$}{
    \textbf{Extract input \& target sequence pairs:} \\
    \(\{\boldsymbol{\eta}_{t},\;
        \boldsymbol{\eta}_{t+1}\}\)

    \textbf{Compute LSTM output:} \\
    \(\boldsymbol{\tilde \eta}_{t+1}
     = f(\boldsymbol{\eta}_{t})\)

    \textbf{Compute loss:} \\
    \(\displaystyle \mathcal{L}_t(\theta_f) \;=\;
\bigl\| \boldsymbol{\tilde \eta}_{t+1} - \boldsymbol{\eta}_{t+1} \bigr\|_2^2
    \)

    \textbf{Update} $\theta_f$ \emph{(Adam optimizer)}
  }
}

\BlankLine
\textbf{Procedure: Training Sequential Predictive Model}

\For{epoch in $1 \dots N_{\text{epoch}}$}{
  \If{(epoch \(\bmod\) $N_{\text{UI}}$) == 0}{\Comment*[r]{Generate or Update Supplementary Dataset}
    \For{$t$ in $1 \dots \bigl(\text{length}(\mathcal{D}_1) - \delta \bigr)$}{
      \textbf{Apply trained model to generate a future state:} \\
      \(\displaystyle
        \tilde{\boldsymbol{\eta}}_{t+\delta}
        \;=\;
        f^\delta(
          \boldsymbol{\eta}_{t }
        )
      \)

      \textbf{Append predicted sequence to} $\mathcal{D}_{\delta}$:
      \[
        \mathcal{D}_{\delta} \;\leftarrow\;
          \mathcal{D}_{\delta} \;\cup\;
            \Bigl\{\,\tilde{\boldsymbol{\eta}}_{t+\delta}\Bigr\}
      \]
    }

    \textbf{Update combined dataset:}\\
    \(\mathcal{D} \;=\; \mathcal{D}_1 \;\oplus\; \mathcal{D}_{\delta}\)

  }
  \For{$t$ in $1 \dots \text{length}(\mathcal{D})$}{\Comment*[r]{Train Predictive Model with Updated Dataset}
    \textbf{Extract input \& target by Eq.~\eqref{eq:acqfun}:} \\
    \(\{\boldsymbol{\eta}^I_{t },\;
        \boldsymbol{\eta}_{t+1}\} \)

    \textbf{Compute output:}\\
    \(\tilde{\boldsymbol{\eta}}_{t+1}
     = f(\boldsymbol{\eta}^I_{t })
    \)

    \textbf{Compute loss:}\\
    \(\displaystyle
      \mathcal{L}_t(\theta_f) \;=\;
      \gamma_i \,\bigl\|\boldsymbol{\tilde \eta}_{t+1} - \boldsymbol{\eta}_{t+1}\bigr\|_2^2
    \)

    \textbf{Update} $\theta_f$ \emph{(Adam optimizer)}
  }
}

\end{algorithm}
\end{minipage}

In addition, an instance of \ac{SPF} with $\delta=3$ is provided in Fig.~\ref{fig:frameworks_compar}.

\section{\hao{Numerical Example: 2D Burgers' Equation}}
\label{sec:burger}

\hao{In the preceding section, we compared the SPF with traditional one-step-ahead prediction. To evaluate and compare their performances, we will employ the Burgers' equation benchmark case. }

\hao{Burgers' equation is a fundamental PDE occurring in various areas, such as fluid mechanics, nonlinear acoustics, and gas dynamics. In our evaluation of the SPF, we employ the 2D Burgers' equation problem. The domain for the simulation is defined as a 128×128 grid. The boundaries of these squares are configured with Dirichlet boundary conditions. The viscosity is 0.01$N \cdot s \cdot m^{-2}$ and the initial velocity ranges from 1.5$m\cdot s^{-1}$ to 5$m\cdot s^{-1}$. We simulated and stored both velocity components $u$ and $v$ at each spatial location, resulting in a final data shape of $128\times128\times2$. The equations are presented as:}

\begin{gather}
\label{eq:burgers}
    \frac{\partial u}{\partial t}+u\frac{\partial u}{\partial x} +v\frac{\partial u}{\partial y} = \frac{1}{Re}(\frac{\partial^2 u}{\partial x^2} + \frac{\partial^2 u}{\partial y^2}) \notag \\
    \frac{\partial v}{\partial t}+u\frac{\partial v}{\partial x} +v\frac{\partial v}{\partial y} = \frac{1}{Re}(\frac{\partial^2 v}{\partial x^2} + \frac{\partial^2 v}{\partial y^2})
\end{gather}
\hao{where $u$ and $v$ represent the velocity components and $t$ is time, $x$ and $y$ represent the coordinate system. $Re$ is the Reynolds number, which can be calculated by $Re = \frac{VL}{\upsilon}$, where $V$ is the flow speed, specified as initial velocity, $L$ is characteristic linear dimension and $\upsilon$ is viscosity.}

\subsection{Performance of Convolutional Autoencoder Model}

\begin{figure}
    \centering
    \includegraphics[width=1\linewidth]{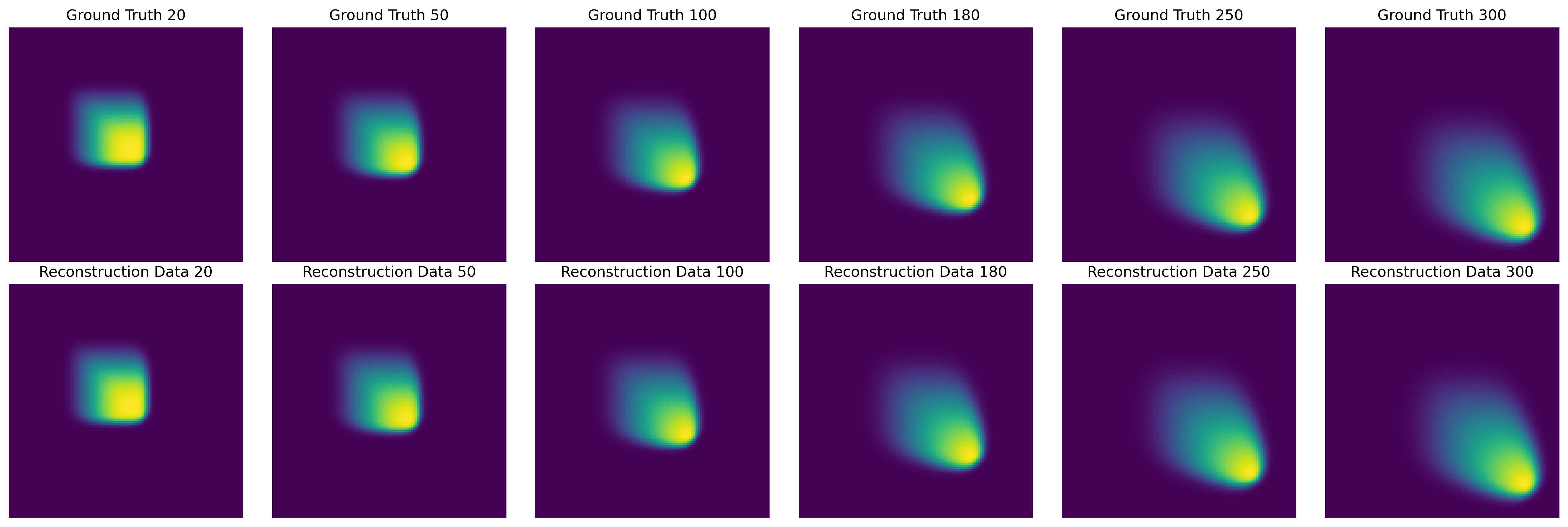}
    \caption{\hao{Illustration of the performance of CAE in the test dataset at different time steps for Burgers' equation}}
    \label{fig:burger_cae}
\end{figure}

\hao{In this study, a \ac{CAE} is employed to compress the solution fields of the 2D Burgers’ equation into a 512-dimensional latent space. The model is trained on 100 simulations and evaluated on 50 unseen simulations, with each simulation consisting of 300 time steps. These simulations differ by their initial conditions, where the initial velocity fields vary within a range of 1.5$m \cdot s^{-1}$ to 5$m \cdot s^{-1}$ following a uniform distribution. To thoroughly evaluate the performance of SPF in long-term prediction, we intentionally maintain a relatively high-dimensional latent space to preserve as much structural and dynamic information as possible during compression. The reconstruction quality is illustrated in Fig.~\ref{fig:burger_cae}.}

\subsection{Long-Term Prediction Performance}

\begin{figure}[h!]
    \centering
    \begin{minipage}{1.\textwidth}
        \begin{subfigure}[t]{0.8\textwidth} 
            \centering
            \includegraphics[width=1.\textwidth]{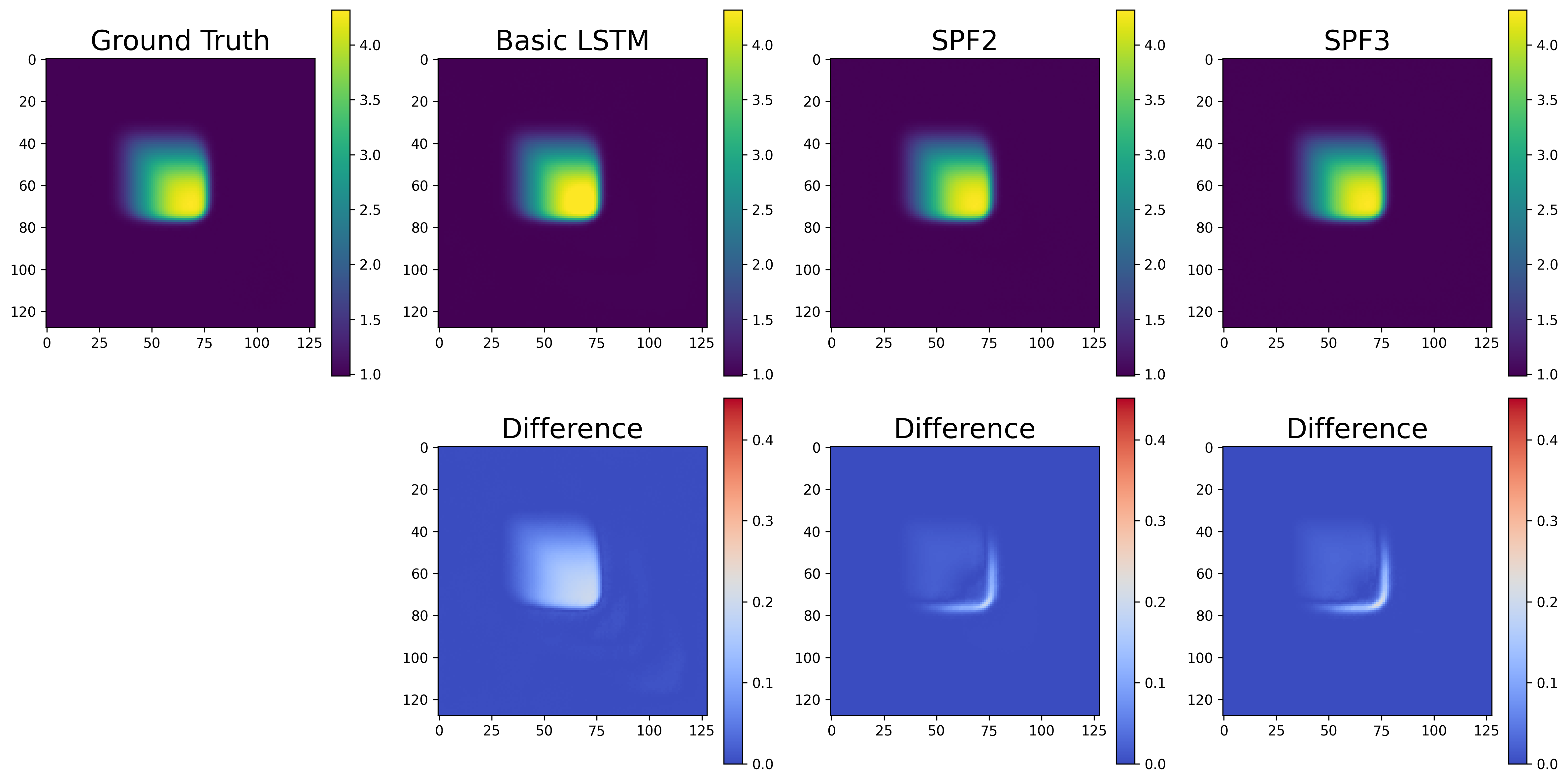}
            \caption{Step 24 Prediction Results}
            \label{fig:Burgerstep24fulldataset}
        \end{subfigure}

        \vspace{1em} 

        \begin{subfigure}[t]{0.8\textwidth}
            \centering
            \includegraphics[width=1.0\textwidth]{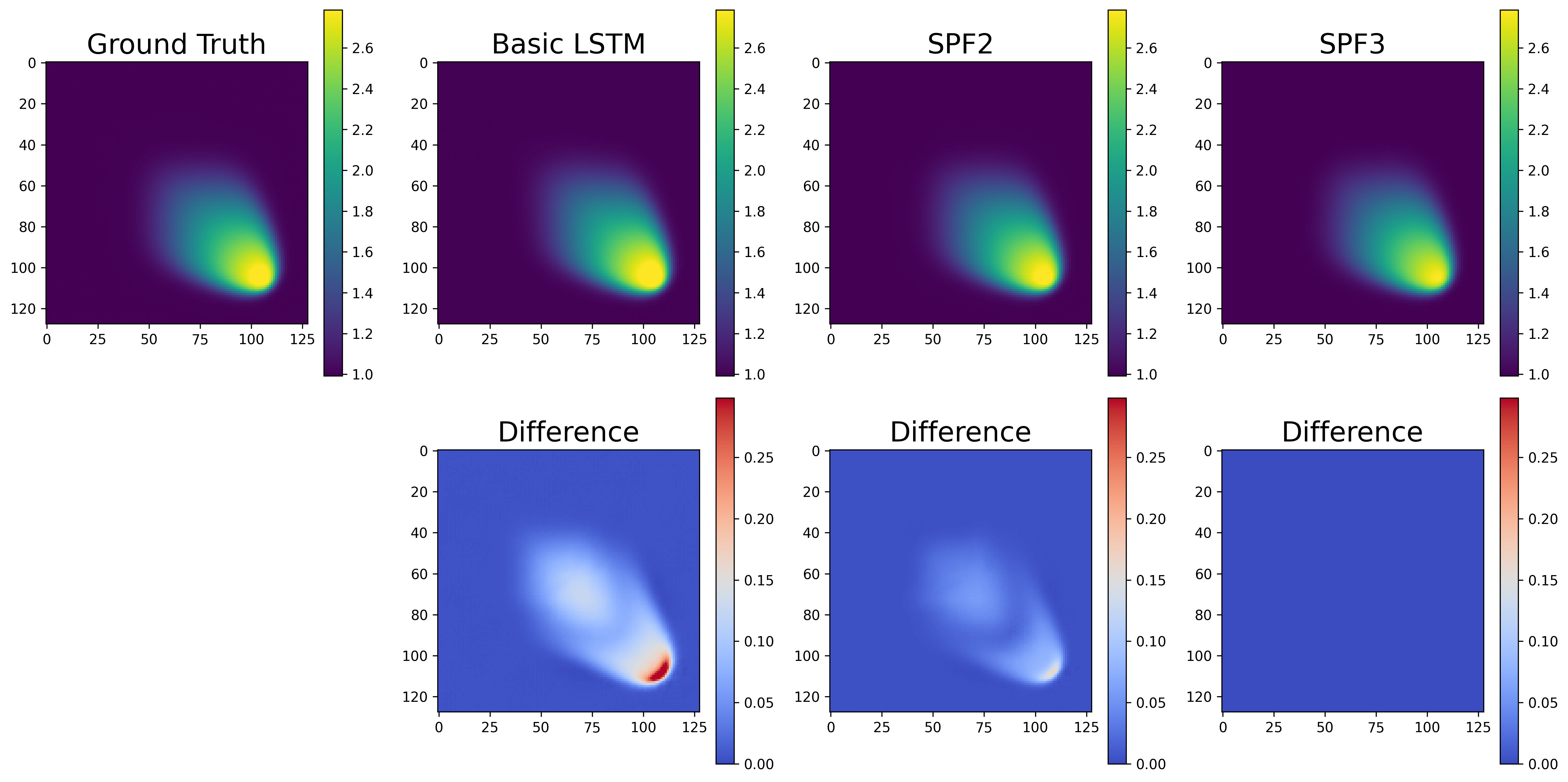}
            \caption{Step 270 Prediction Results}
            \label{fig:Burgerstep270fulldataset}
        \end{subfigure}
    \end{minipage}
    \caption{\hao{Prediction results ($u$ dimension) and difference with ground truth across different models for Burgers' equation.}}
    \label{fig:burger_snapshots}
\end{figure}

\hao{To assess the long-term prediction capability of the proposed \ac{SPF}, we evaluate and compare its performance against a baseline LSTM model using accumulated error and \ac{SSIM} over multiple forecasting steps. The SPF and baseline LSTM share the same \ac{CAE}. The predicted velocity fields and their differences from the ground truth at both early (step 24) and late (step 270) prediction stages are visualized in Fig.~\ref{fig:burger_snapshots}.} 

\begin{figure}
    \centering
    \includegraphics[width=1\linewidth]{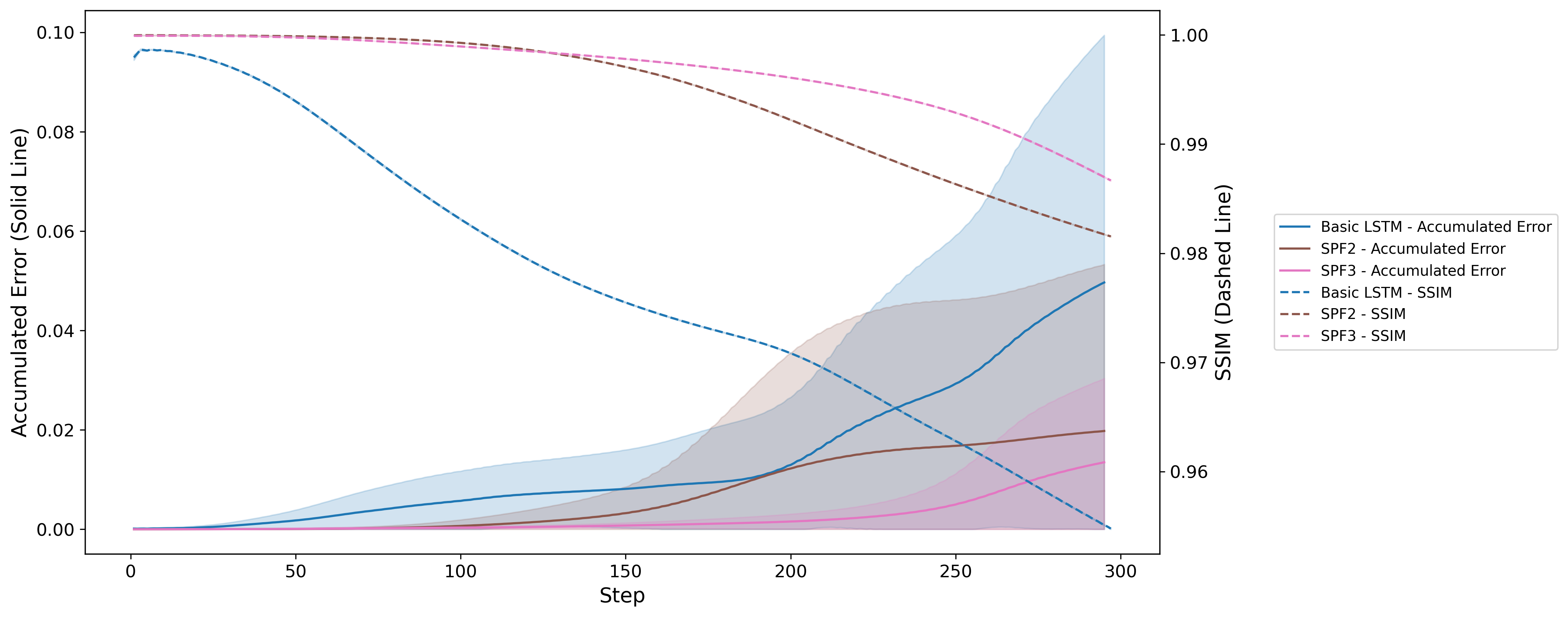}
    \caption{\hao{Long-term prediction performance comparison between baseline LSTM and SPF variants on the Burgers' equation. Solid lines represent accumulated error; dashed lines represent SSIM.}}
    \label{fig:burger_longterm}
\end{figure}

\hao{In Fig.~\ref{fig:burger_longterm}, solid lines represent the accumulated error over prediction steps, while dashed lines denote the SSIM values. The accumulated error is calculated as the cumulative \ac{MSE} between the predicted and ground truth latent representations at each stepthe, which reflects the MSE propagated through autoregressive predictions over time. The SSIM metric evaluates the structural similarity between the predicted and true solution fields, offering a perceptual measure of fidelity.}

\hao{We observe that the baseline \ac{LSTM} model suffers from noticeable error accumulation as the prediction horizon increases. Its SSIM score also deteriorates significantly beyond 100 steps, indicating a loss of spatial consistency in the predicted solutions. In contrast, SPF variants (e.g., SPF2 and SPF3) exhibit substantially lower accumulated error and maintain high SSIM values throughout the long-term horizon. Among them, SPF3 demonstrates the best performance, achieving the lowest error and highest structural consistency.}

\hao{These findings highlight the effectiveness of the SPF in stabilizing multi-step rollouts by leveraging its mechanism and design, which mitigates compounding errors over time.} 

\section{Numerical Example: Shallow Water}
\label{sec:sw}

\hao{To further assess the performance of the proposed \ac{SPF}, we compare it with the existing \ac{ATF} and \ac{PF} frameworks. A shallow water case study is used to benchmark their effectiveness. In this comparison, both \ac{ATF} and \ac{PF} serve as reference methods against which the efficacy of the \ac{SPF} is evaluated. Additionally, we also evaluate the SPF’s ability to incorporate physical constraints.}

The shallow water equations are a set of hyperbolic partial differential equations that describe the flow below a pressure surface in a fluid, typically water. The governing equations are:

\begin{align}
\frac{\partial h}{\partial t} + \frac{\partial (hu)}{\partial x} + \frac{\partial (hv)}{\partial y} &= 0 \notag \\
\frac{\partial (hu)}{\partial t} + \frac{\partial (hu^2 + \frac{1}{2} g h^2)}{\partial x} + \frac{\partial (huv)}{\partial y} &= 0 \notag \\
\frac{\partial (hv)}{\partial t} + \frac{\partial (huv)}{\partial x} + \frac{\partial (hv^2 + \frac{1}{2} g h^2)}{\partial y} &= 0
\label{eq:Shallow_Water_Equation}
\end{align}

where $h$ is the total water depth (including the undisturbed water depth) with units of meters ($m$), $u$ and $v$ are the velocity components in the x (horizontal) and y (vertical) directions with units of meters per second ($m/s$), respectively, and $g$ is the gravitational acceleration, typically measured in meters per second squared ($m/s^2$). For our simulations, the numerical results are obtained by solving the shallow water equations using a combination of the finite difference method for spatial discretization and the Euler method for time integration. The simulation operates on a \(64 \times 64\) grid, which incorporates three channels representing three variables, the velocity components $u$, $v$, and the water height $h$. The initial conditions introduce a cylindrical disturbance in the water height, where the central cylinder varies in height from 0.2 to 1$m$ and in radius from 4 to 16 grid units. This setup permits an extensive analysis of wave dynamics and fluid behavior. The water depth outside the disturbance remains constant at 1$m$. \hao{The dataset comprises three independently generated subsets: 300 simulations of data used exclusively for training, 100 simulations for validation, and 100 simulations reserved strictly for the final testing phase. These simulations differ in their initial conditions, consisting of the initial radius and water height of the central cylindrical region following a uniform distribution. For each dataset, we analyze 300 timesteps, starting from timestep 51 to 350, thereby excluding the initial 50 timesteps to avoid the transient effects. Specifically, during the testing phase, model inputs are not from the initial timesteps (51–53) but are explicitly selected from timesteps 201 to 203, ensuring an accurate evaluation of the model’s predictive capability for realistic scenarios beyond the initial transient period.}

\subsection{Performance of Convolutional Autoencoder Model}

\begin{figure}
    \centering
    \includegraphics[width=1\linewidth]{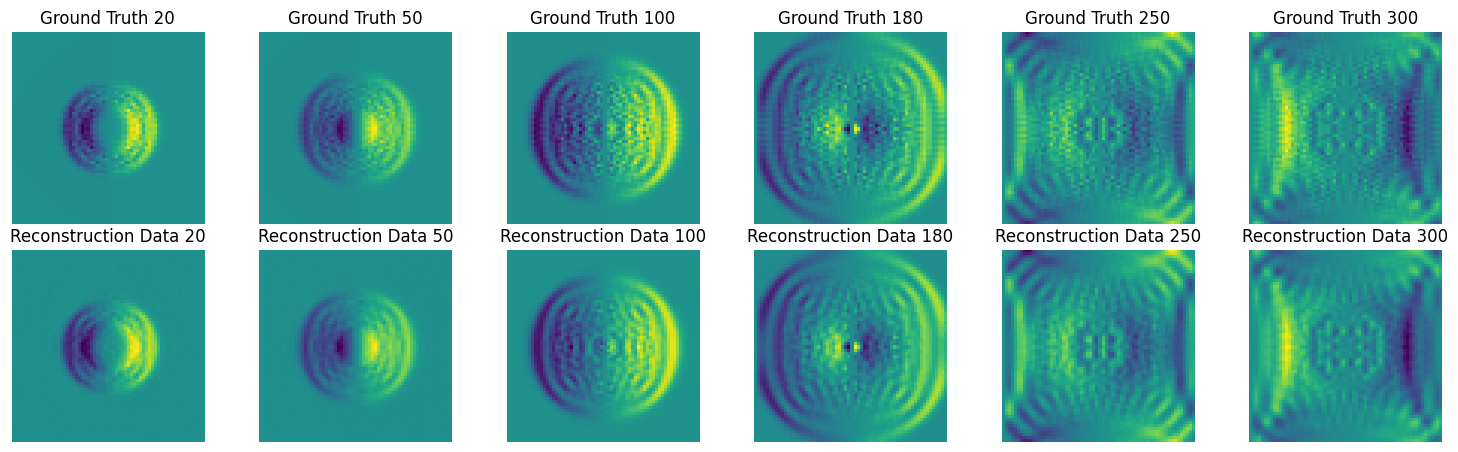}
    \caption{Illustration of the performance of CAE in the test dataset at different time steps}
    \label{fig:CAE_performance}
\end{figure}

\hao{Similar to the Burgers' equation, a \ac{CAE} is employed for the shallow water system to compress the spatiotemporal data into a 512-dimensional latent space. The reconstruction performance is presented in Fig.~\ref{fig:CAE_performance}.}

\subsection{Long-Term Prediction Performance}

Obtaining the compressed data $\eta$ from \ac{ROM}, \ac{LSTM} is used to play the role as surrogate model for latent dynamics~\cite{hochreiter1997long}. Our evaluation involves a sequence of 300 timesteps (from 1 to 300). Specifically, to assess the model's predictive capabilities, timesteps 151 to 153 serve as inputs. We then engage in a sequential 3-to-3 prediction cycle, where the output from the models is recycled back as the input for the subsequent predictions. This iterative process, which shares the same fundamental with \ac{ATF}, allows for an extended analysis over multiple cycles, effectively testing the models' capacity for long-term prediction.

\subsubsection{Performance Comparison between Different Training Frameworks}
Based on the evaluation method, three distinct training framework, \ac{ATF}, \ac{PF} and \ac{SPF}, are employed to explore their ability to handel the long-term predictions. These frameworks, implemented with depths $\delta=2,3$, referred to as ATF2, ATF3, PF2, PF3, SPF2, and SPF3, respectively, are compared against a basic model trained via single-step-ahead predictions. 

\begin{figure}[h!]
    \centering
    \begin{minipage}{1.1\textwidth}
        \hspace*{-2.5cm} 
        \begin{subfigure}[t]{1.\textwidth} 
            \centering
            \includegraphics[width=1.2\textwidth]{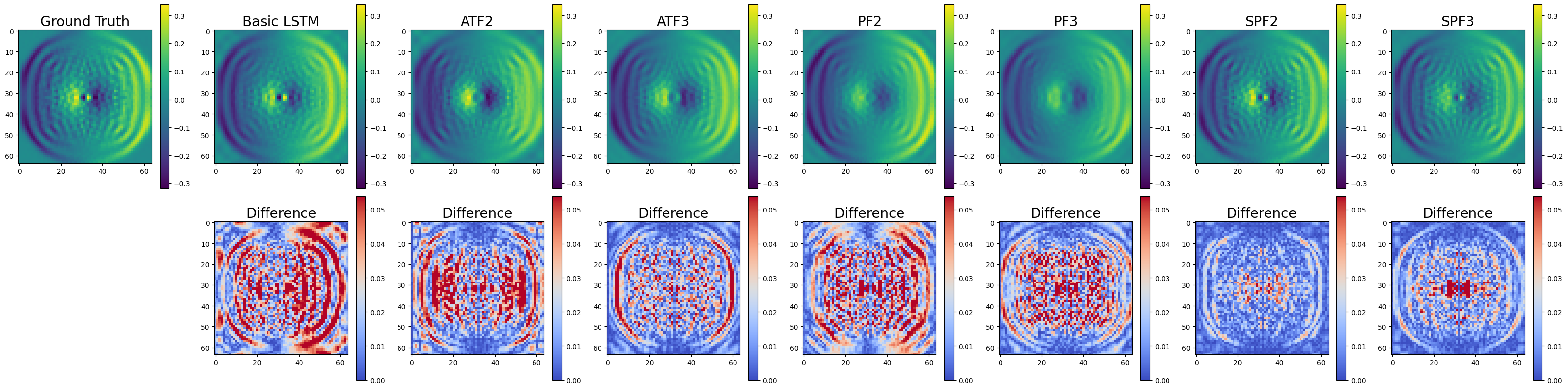}
            \caption{Step 24 Prediction Results}
            \label{fig:step24fulldataset}
        \end{subfigure}

        \vspace{1em} 

        \hspace*{-2.5cm} 
        \begin{subfigure}[t]{1.\textwidth}
            \centering
            \includegraphics[width=1.2\textwidth]{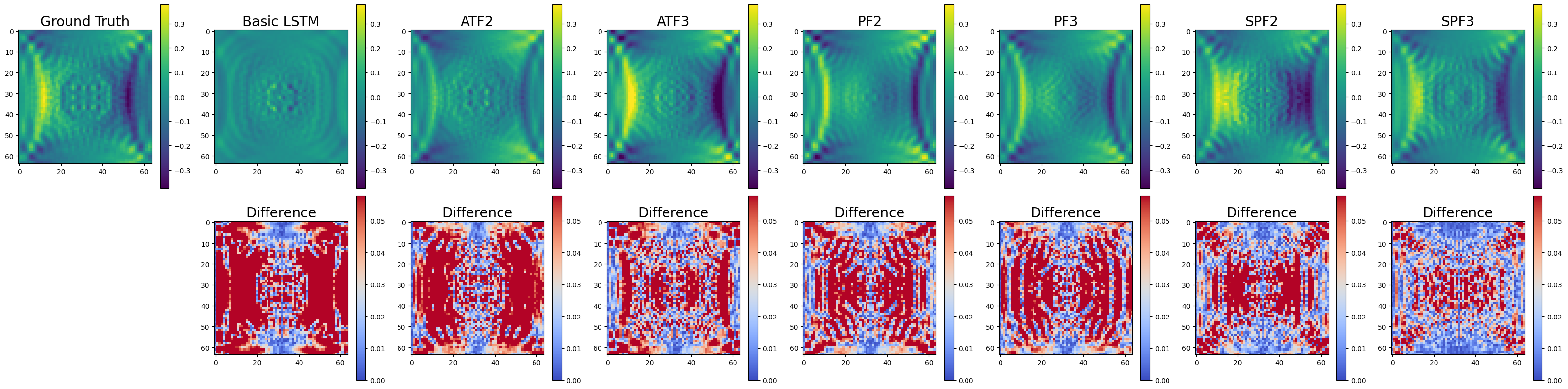}
            \caption{Step 140 Prediction Results}
            \label{fig:step140fulldataset}
        \end{subfigure}
    \end{minipage}
    \caption{Prediction results ($u$ dimension) and difference with ground truth across different models for shallow water system.}
    \label{fig:long-term prediction}
\end{figure}

The performance of these models for short-term (step=24) and long-term (step=140) predictions is illustrated in Fig.~\ref{fig:step24fulldataset} and Fig.~\ref{fig:step140fulldataset}. The difference in Fig.~\ref{fig:long-term prediction} is calculated at each point as the absolute value of direct subtraction of the predicted value from the actual value, which represents the absolute error at each point. This comparison highlights the relative efficacy of each approach under varying prediction lengths. As shown in Fig.~\ref{fig:step24fulldataset}, all models perform well in short-term predictions, while the basic \ac{LSTM} initially displays asymmetrical behavior. For longer predictions (step=140), the basic \ac{LSTM} exhibits gradient disappearance, resulting in a flattened prediction, while the other models maintain a more robust performance. Among these, all of \ac{ATF}, \ac{PF} and \ac{SPF} models with a depth of 3 ($\delta=3$) outperform their $\delta=2$ counterparts, with \ac{SPF} models consistently surpassing \ac{ATF} and \ac{PF} models in terms of accuracy across both prediction lengths.




\begin{figure}
    \centering
    \includegraphics[width=1\linewidth]{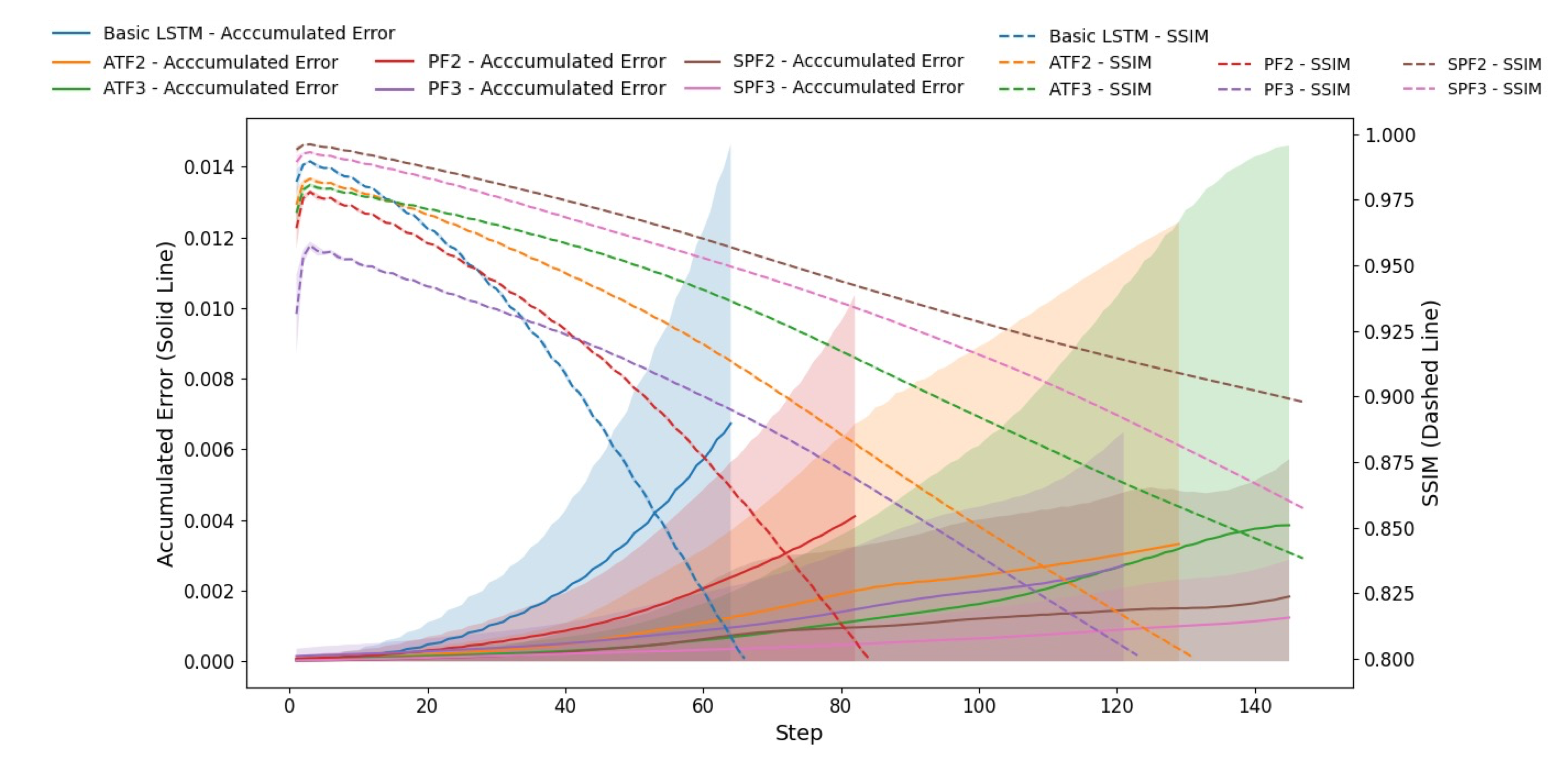}
    \caption{Performances comparison across different models}
    \label{fig:fulldataset_performance}
\end{figure}

The quantified results are illustrated in Fig.~\ref{fig:fulldataset_performance} where two different metrics, \ac{MSE} and \ac{SSIM}, are used to assess the models' performance. In this study, if \ac{SSIM} falls below 0.8, the subsequent data points are not plotted, as the results below the threshold are deemed non-informative. As illustrated in Fig.~\ref{fig:fulldataset_performance}, the accumulated error and \ac{SSIM} of all models deteriorate across the time steps. The basic \ac{LSTM} shows a rapid increase in accumulated error and a significant drop in \ac{SSIM}. In contrast, the increase in accumulated error and decrease in \ac{SSIM} for the other models is relatively gentle, with models at $\delta=3$ outperforming those at $\delta=2$. Notably, the \ac{SPF} models demonstrate less error and uncertainty compared to \ac{ATF} and \ac{PF}, indicating more reliable and consistent predictions, aligning with the results shown in Fig.~\ref{fig:long-term prediction}. Interestingly, in the initial steps, the basic \ac{LSTM} outperforms the models from the \ac{ATF} and \ac{PF}, illustrating a trade-off in these two models between short-term accuracy and long-term prediction capabilities. In contrast, the \ac{SPF} manages both short-term and long-term predictions more effectively.

\subsubsection{Energy Consistency}

In the shallow water system, energy conservation is fundamental because the governing equations inherently ensure that, in the absence of external forces or dissipation, the total energy—comprising kinetic energy from fluid motion and potential energy from water height—remains constant over time. This makes energy conservation a natural and reliable metric for evaluating the physical consistency of predictive models.

\begin{figure}
    \centering
    \includegraphics[width=1\linewidth]{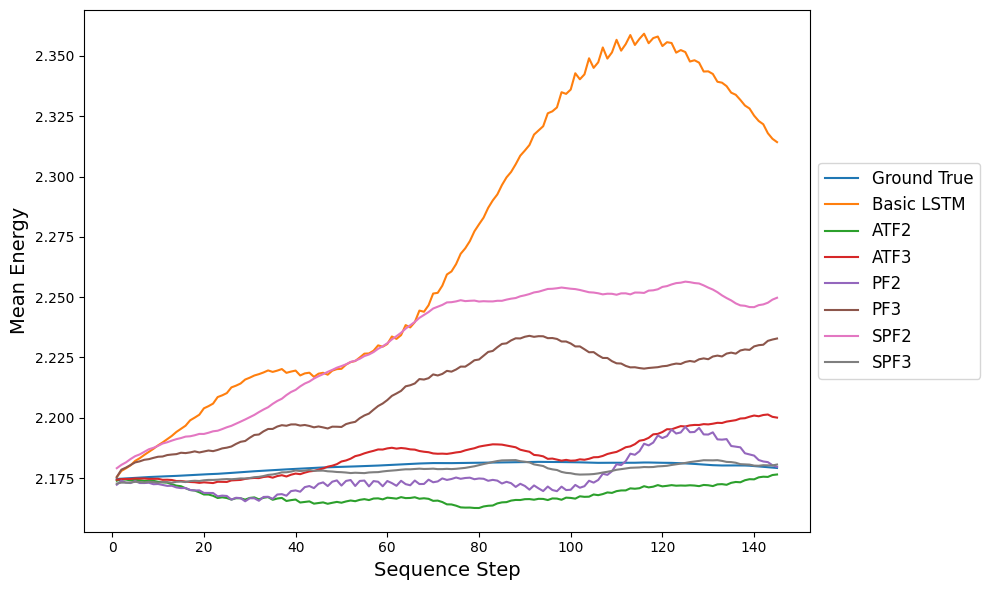}
    \caption{Energy consistency comparison across different models}
    \label{fig:energy_consistency_fulldataset}
\end{figure}

As shown in Fig.~\ref{fig:energy_consistency_fulldataset}, the basic \ac{LSTM} model demonstrates a significant deviation from the ground truth, reflecting its inability to maintain energy consistency over time. The \ac{ATF} models, which leverage full-step backpropagation, exhibit relatively stable energy levels due to their improved handling of short-term errors. In contrast, the performance of \ac{PF} models deteriorates as $\delta$ increases, with error accumulation becoming more pronounced at larger $\delta$. The \ac{SPF}3 model, depicted by the grey curve, aligns closely with the ground truth, showcasing its ability to effectively mitigate error accumulation and achieve a balance between short-term accuracy and long-term predictive performance, while preserving energy conservation.

\hao{
Building upon the importance of energy consistency discussed above, we further enhance our algorithm by embedding energy conservation constraints into the training framework, denoted as \ac{SPF}2 with Physics Constraint (\ac{SPF}2-PC). To enforce this constraint, we introduce a physics-based regularisation term that penalises deviations in the total system energy over time. The modified loss function is defined as:
\begin{equation}
\mathcal{L}_{\text{SPF2-PC}} = \mathcal{L}_{\text{MSE}} + \lambda_{\text{PC}} \cdot \left| E_{\text{pred}} - E_{\text{true}} \right|,
\end{equation}
where $\mathcal{L}_{\text{MSE}}$ is the standard prediction loss, $\lambda_{\text{PC}}$ is a weighting coefficient, which is set as 1 in this test, and $E_{\text{pred}}, E_{\text{true}}$ denote the total predicted and true energy respectively, computed as
\begin{equation}
E = \int \left( \frac{1}{2}(u^2 + v^2) + \frac{1}{2}g h^2 \right) \, dx\, dy,
\end{equation}
where $u, v$ are the horizontal velocity components, $h$ is the fluid height, and $g$ is gravitational acceleration.}

\begin{figure}
    \centering
    \includegraphics[width=1.2\linewidth]{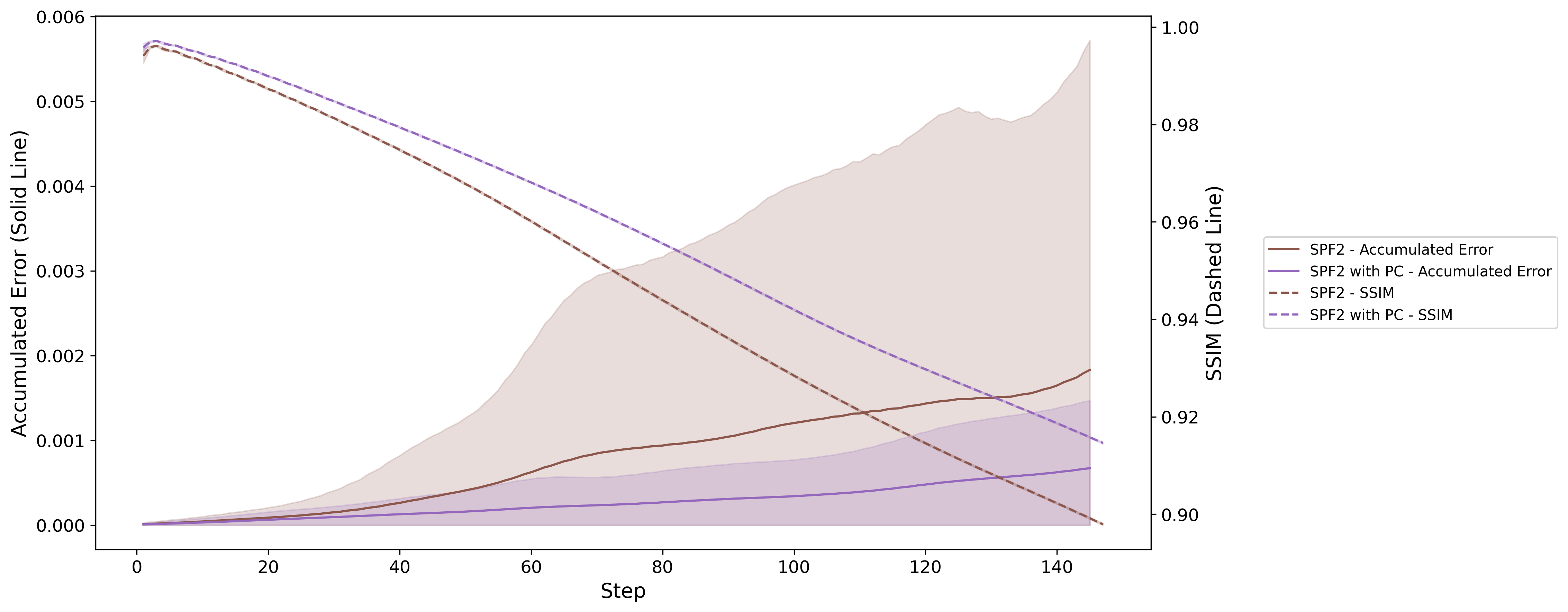}
    \caption{\hao{Comparison of long-term prediction performance between SPF2 and SPF2 with physics constraint (PC).}}
    \label{fig:seq_with_PC}
\end{figure}

\hao{As illustrated in Fig.~\ref{fig:seq_with_PC}, the performance comparison between the original SPF2 model and SPF2-PC highlights the advantages of this approach. The accumulated error (solid lines) for the SPF2-PC remains consistently lower across the entire prediction horizon, indicating enhanced accuracy and robustness. Meanwhile, the \ac{SSIM} (dashed lines) for SPF2-PC consistently maintains higher values compared to the original SPF2, suggesting improved fidelity of the predicted states. 
These results demonstrate that \ac{SPF} can be effectively integrated with physics-informed approaches to further enhance prediction accuracy.
}

\subsection{Model Performance with limited training data}

\begin{figure}
    \hspace*{-2.cm}
    \centering
    \includegraphics[width=1.2\linewidth]{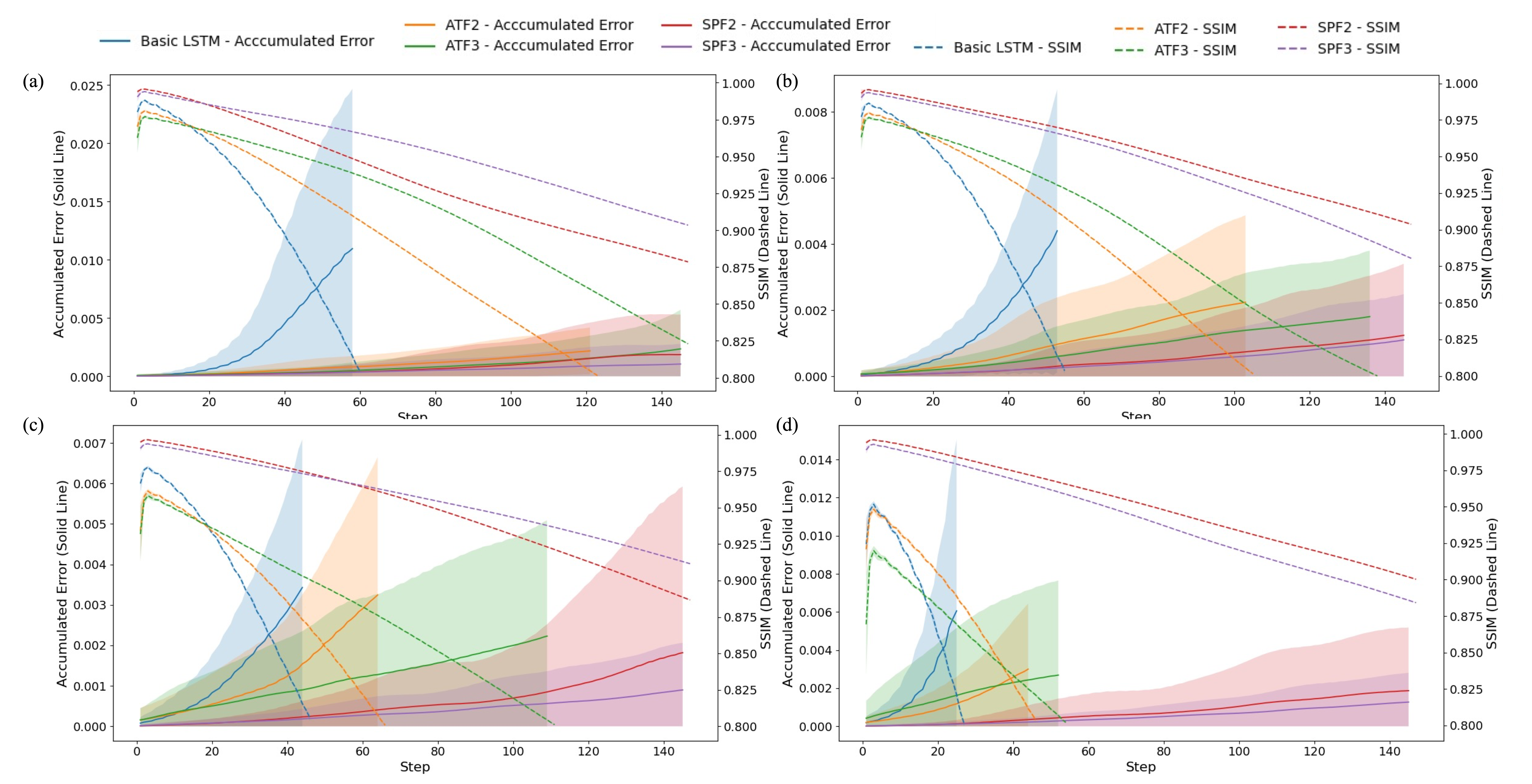}
    \caption{Performances comparison across different models on (a) 50\% (b) 30\% (c) 10\% (d) 5\% limited training data }
    \label{fig:performance_of_partial_data}
\end{figure}

In this section, we assess the adaptability and performance of various models when trained on limited training data to understand their efficacy under conditions of limited data. As previous results indicate that \ac{PF} does not outperform \ac{ATF} in the shallow water case, we therefore concentrate our additional comparisons on \ac{ATF} and \ac{SPF}. As illustrated in Fig.~\ref{fig:performance_of_partial_data}, models were trained using subsets of data comprising 50\%, 30\%, 10\%, and 5\%. These experiments revealed a consistent trend: performance metrics, including accumulated error and \ac{SSIM}, progressively deteriorated as the dataset size decreased. This was particularly noticeable with dataset sizes reduced to 10\% and especially 5\%, where the decline became sharply pronounced. The \ac{ATF} managed to maintain some effectiveness with moderately reduced datasets, yet its performance significantly faltered under severe data constraints found in the 10\% and 5\% conditions. In contrast, the \ac{SPF} showed remarkable resilience, likely benefiting from its ability to utilize supplementary datasets treated as synthetically generated data, which helps to overcome the limitations of sparse datasets. This capability of \ac{SPF} to effectively handle extreme data scarcity suggests its suitability for applications where data is inherently limited.

\begin{figure}
    \centering
    \includegraphics[width=1\linewidth]{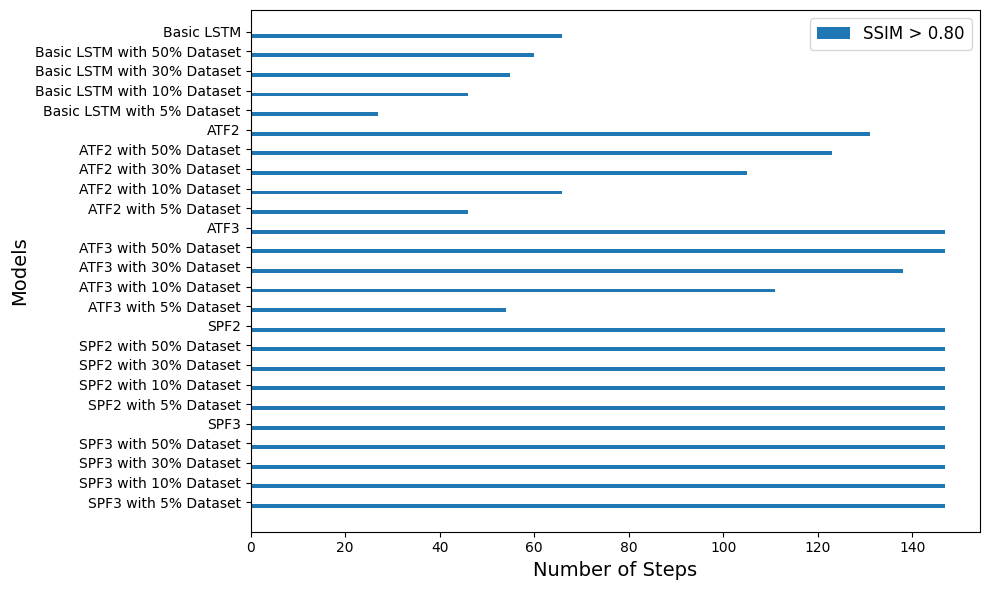}
    \caption{Step count analysis of LSTM, ATF, and SPF models at an SSIM threshold of 0.80 across various dataset sizes}
    \label{fig:steps_counting}
\end{figure}

The bar chart in Fig.~\ref{fig:steps_counting} presents a comprehensive step count analysis where the \ac{SSIM} exceeds a threshold of 0.80, across a range of dataset sizes for three different model types. Each bar represents the number of steps within the test period where the \ac{SSIM} of the respective model remains above 0.80, providing a direct measure of how well each model maintains structural similarity under constrained data conditions. Notably, the performance across models varies significantly with dataset size: models trained with larger datasets consistently achieve higher step counts above the SSIM threshold, demonstrating greater predictive reliability. Conversely, all models exhibit a steep decline in the number of steps meeting the SSIM criterion when trained on datasets as small as 5\%, highlighting the challenges posed by severe data scarcity. This analysis underscores the various capabilities of basic LSTM, ATF, and SPF models in dealing with data limitation, with SPF models generally showing superior performance in maintaining higher structural similarity across more steps, particularly in more constrained datasets.

\subsection{Timestep Extrapolation}

When training a predictive model to forecast the future state of a physical system, the ultimate objective is to enable the model to fully comprehend and embody the underlying physical relationship, such as Eq.~\eqref{eq:Shallow_Water_Equation} in this study. The comprehension ensures the model's independency from the specific dataset used in training, allowing this application in diverse scenarios. To access whether the model meets this criterion, extrapolation tests are conducted. These tests are crucial for evaluating the robustness and generalizability for these frameworks, confirming its ability to provide reliable prediction even when applied to the data or condition beyond its initial training set.

In this paper, an additional set of 100 timesteps, from timestep 351 to 450, is included in the test set for extrapolation purposes. Practically, the final steps in the training phase, specifically timestep 348 to 350, are used as the initial input for the trained models. Then the 3-to-3 predictive cycle are applied to forecast the subsequent timesteps. Additionally, as the \ac{CAE} is trained by training set, spanning timestep from 51 to 350, the encoding-decoding process inherently introduces some noise into the finial output. Therefore, the results of pure \ac{CAE} are also presented in the subsequent results comparison to illustrate its effect.

\begin{figure}
    \centering
    \begin{subfigure}[t]{\linewidth}
        \includegraphics[width=\linewidth]{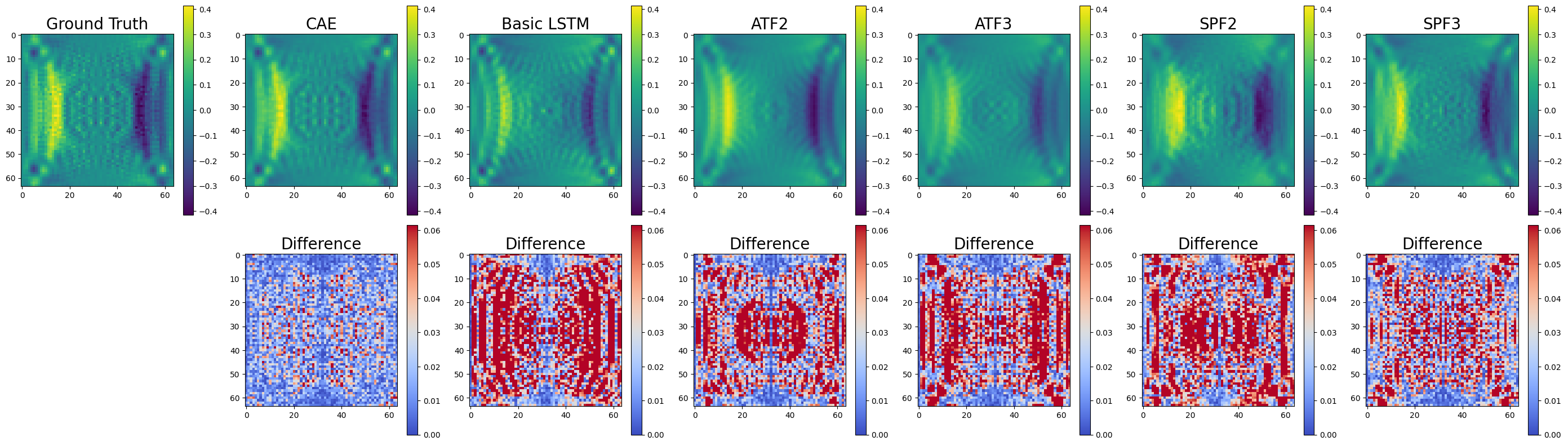}
        \caption{Step 24 Prediction Results for Extrapolation}
        \label{fig:extra_step24}
    \end{subfigure}

    \begin{subfigure}[t]{\linewidth}
        \includegraphics[width=\linewidth]{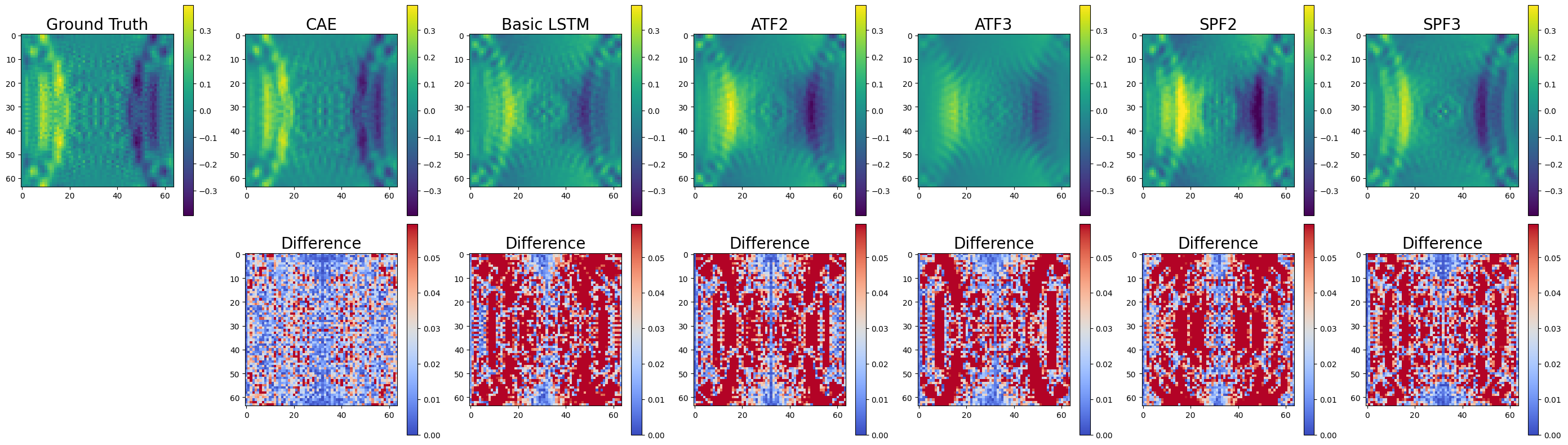}
        \caption{Step 50 Prediction Results for Extrapolation}
        \label{fig:extra_step50}
    \end{subfigure}
    \caption{Prediction results ($u$ dimension) and difference with ground truth across different models for extrapolation}
    \label{fig:extrapolation prediction}
\end{figure}

\begin{figure}
    \centering
    \includegraphics[width=1\linewidth]{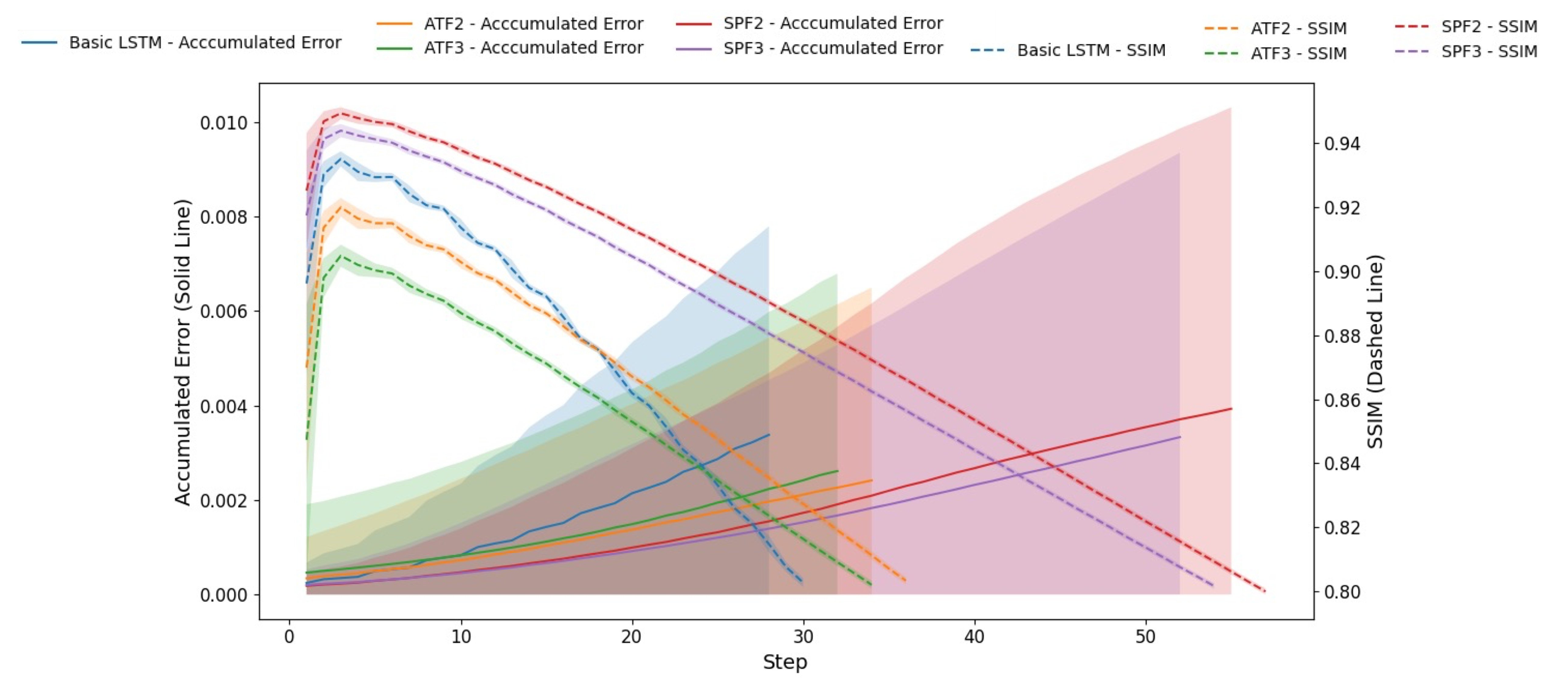}
    \caption{Performances comparison across different models on extrapolation prediction}
    \label{fig:extra_compar}
\end{figure}

The extrapolation prediction results at Step 24 and 50 are illustrated in Fig.~\ref{fig:extrapolation prediction} and the quantified data are shown in Fig.~\ref{fig:extra_compar}. From both of these figures, there is no huge discrepancy between different predictive results of models compared with previous cases. However, when we focus on the \ac{SSIM} in Fig~\ref{fig:extra_compar}, it can be seen that \ac{SPF} outperforms the other models and the \ac{ATF} performs the worst at the initial stage.

\subsection{Robustness Evaluation with Noisy Data}

In practical applications, especially when dealing with complex systems, models frequently encounter data that is tainted with noise. This noise may arise from various sources such as inaccurate measurements, inherent uncertainties in the system, or external interferences. It is critically important to ensure that models intended for complex physical systems maintain their robustness and predictive accuracy despite these noisy conditions. In order to further evaluate the stability of \ac{SPF} in this particular environment, we conduct a noisy experiment within the shallow water systems. By utilizing a model that is trained on data without any noise, we conduct an evaluation of its capacity to make accurate predictions on a dataset that is intentionally contaminated with synthetic noise. This simulation aims to replicate the obstacles encountered in real-world scenarios.

In our experiments, to ensure the representativeness of numerical tests, we utilize spatial correlation patterns that are both homogeneous and isotropic with respect to the spatial Euclidean distance $r=\sqrt{\Delta_x^2+\Delta_y^2}$. This means that they remain unchanged under rotations and translations. We employ these correlation patterns to simulate data errors stemming from various sources. In this context, we consider a Matern type of correlation function:

\begin{equation}
    \epsilon(r) = (1+\frac{r}{L})e^{(-\frac{r}{L})}
\end{equation}
where $L$ is the typical correlation length scale. For simplicity, we set $L = 4$ in this study.

\begin{figure}
\hspace*{-2.cm}
    \centering
    \includegraphics[width=1.2\linewidth]{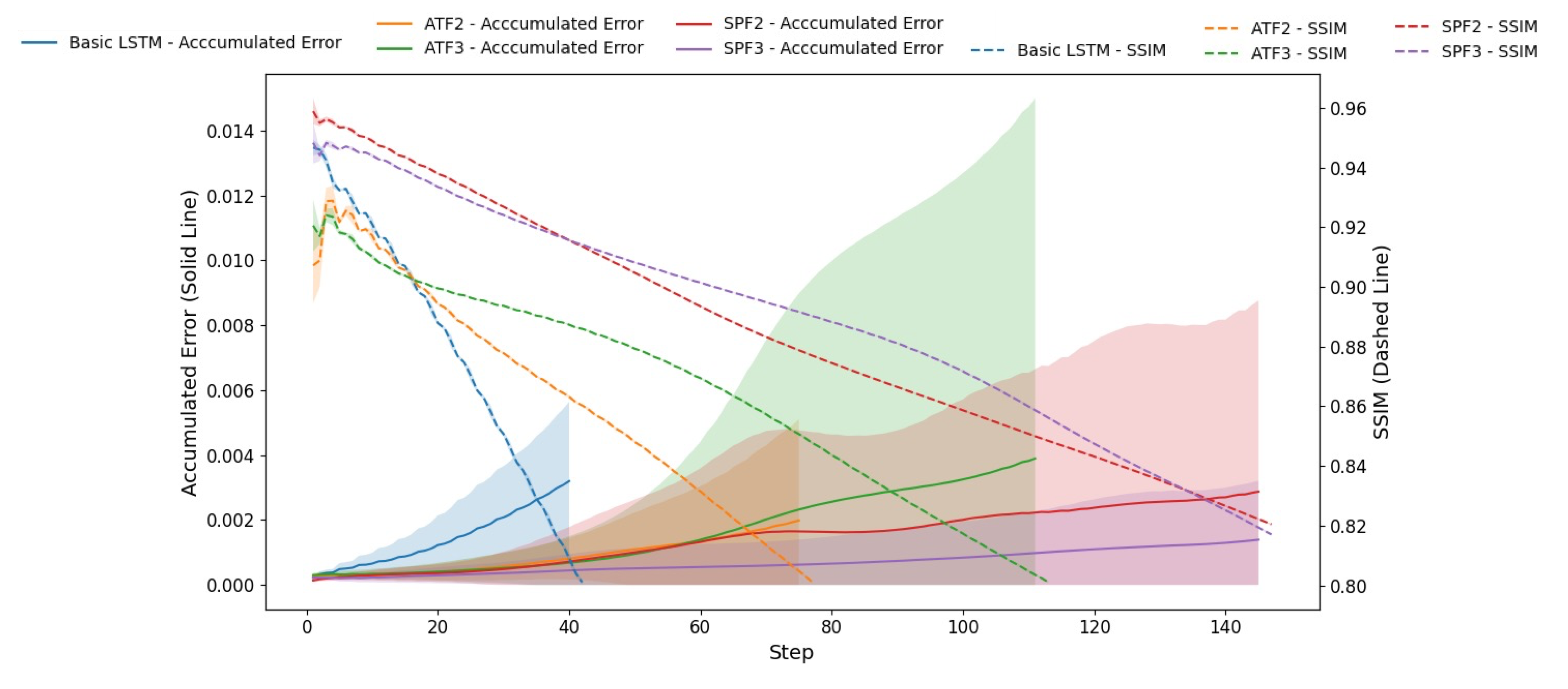}
    \caption{Performances comparison across different models on noisy data}
    \label{fig:noisy_performance}
\end{figure}

In the noise simulation, we add noise to the initial input data by scaling it with a specified coefficient, resulting in noisy data for evaluation. This noisy data is then fed into all trained models for iterative predictions, as described in Eq.~\eqref{eq:LSTM_Predict_alt}, where the $\eta_0$ is replaced by noisy data. The outcomes depicted in Fig.\ref{fig:noisy_performance} illustrate the models' performances in the presence of noisy data. When compared with the results in Fig.\ref{fig:fulldataset_performance}, it is evident that the basic LSTM model is particularly vulnerable to noise, exhibiting a significant drop in \ac{SSIM}. Both the \ac{ATF} and the \ac{SPF} demonstrate enhanced robustness against noise, with noticeably better \ac{SSIM} scores across most steps. This improvement suggests that the additional processing layers inherent in ATF and SPF may be effective in mitigating the impact of noise on prediction accuracy. Particularly, models with greater depth $\delta$, such as ATF3 and SPF3, show a beneficial effect in noisy environments. While ATF3 occasionally achieves higher \ac{SSIM} values than SPF2 and SPF3, it exhibits a larger standard deviation, indicating less stability compared to SPF3. SPF3, in contrast, tends to maintain higher and more consistent \ac{SSIM} values, indicating its superior ability to preserve structural integrity under noisy conditions, with more predictable performance.

\hao{\subsection{Hyperparameters' Effect in \ac{SPF}}}

\hao{The results above demonstrate the ability of \ac{SPF}. This section investigates the impact of different hyperparameters, specifically probability $p$ and coefficient $\alpha$, on the performance of \ac{SPF} in short-term and long-term prediction scenarios when the model depth $\delta$ is set to 2. The \ac{MSE} and \ac{SSIM} metrics are employed to evaluate performance.}

\hao{Table~\ref{tab:short-term} illustrates the performance of short-term prediction (1 step) for various combinations of hyperparameters $p$ and $\alpha$. Observations indicate that the \ac{SPF} model consistently achieves negligible MSE values across all hyperparameter combinations, demonstrating strong short-term predictive capabilities by applying \ac{SPF}. \ac{SSIM} scores remain notably high (above 0.98), suggesting excellent image reconstruction quality. }

\hao{Table~\ref{tab:long-term} provides the long-term prediction results (150 steps), highlighting more varied outcomes in comparison to short-term predictions. The MSE values across different hyperparameters range between 0.0011 and 0.0028, indicating acceptable but more challenging long-term forecasting. SSIM scores demonstrate substantial variation, ranging between 0.8161 and 0.8959. Optimal performance in terms of MSE (0.0011) occurs at $p=0.5, \alpha=1.0$, whereas the best SSIM (0.8959) occurs at $p=0.5, \alpha=0.75$. These results imply a balanced selection of hyperparameters is crucial for robust long-term predictive performance.}

\hao{Overall, the analysis emphasizes the importance of carefully selecting hyperparameters based on the forecasting horizon, as the optimal set of parameters may also depend on the depth of the SPF and the specific physics problem under consideration.}

\begin{table}[]
\centering
\caption{\hao{Short-term prediction (1 step) performance for various hyperparameter combinations (MSE and SSIM)}}
\label{tab:short-term}
{
\begin{tabular}{ccccc}

\hline
\multicolumn{1}{l}{}    & Metric & $p$=0.25 & $p$=0.5  & $p$=0.75 \\ \hline
\multirow{2}{*}{$\alpha$=0.25} & MSE  & $1.7\times10^{-5}$& $1.5\times10^{-5}$& $1.2\times10^{-5}$\\
                        & SSIM & 0.9937 & 0.9942 & 0.9956 \\
\multirow{2}{*}{$\alpha$=0.5}  & MSE  & $2.6\times10^{-5}$& $1.8\times10^{-5}$& $1.4\times10^{-5}$\\
                        & SSIM & 0.9910 & 0.9937 & 0.9953 \\
\multirow{2}{*}{$\alpha$=0.75} & MSE  & $3.1\times10^{-5}$& $2.0\times10^{-5}$& $1.2\times10^{-5}$\\
                        & SSIM & 0.9893 & 0.9933 & 0.9958 \\
\multirow{2}{*}{$\alpha$=1.0}  & MSE  & $3.9\times10^{-5}$& $2.2\times10^{-5}$& $1.3\times10^{-5}$\\
                        & SSIM & 0.9864 & 0.9924 & 0.9954 \\ \hline

\end{tabular}}
\end{table}

\begin{table}[]
\centering
\caption{\hao{Long-term prediction (150 steps) performance for various hyperparameter combinations (MSE and SSIM)}}
\label{tab:long-term}
{
\begin{tabular}{ccccc}
\hline
\multicolumn{1}{l}{}    & Metric & $p$=0.25 & $p$=0.5  & $p$=0.75 \\ \hline
\multirow{2}{*}{$\alpha$=0.25} & MSE  & 0.0017 & 0.0022 & 0.0025 \\
                        & SSIM & 0.8817 & 0.8524 & 0.8161 \\
\multirow{2}{*}{$\alpha$=0.5}  & MSE  & 0.0021 & 0.0014 & 0.0028 \\
                        & SSIM & 0.8933 & 0.8804 & 0.8379 \\
\multirow{2}{*}{$\alpha$=0.75} & MSE  & 0.0013 & 0.0014 & 0.0015 \\
                        & SSIM & 0.8625 & 0.8959 & 0.8903 \\
\multirow{2}{*}{$\alpha$=1.0}  & MSE  & 0.0016 & 0.0011 & 0.0020 \\
                        & SSIM & 0.8504 & 0.8867 & 0.8441 \\ \hline
\end{tabular}}
\end{table}

\subsection{Computational Cost}

In this study, we utilize the GPU Tesla T4 with the \textit{Google Colab} computing platform to optimize storage and computational costs. Compressed data from a \ac{CAE} are employed to train various LSTM training frameworks directly. This ensures that memory usage remains small across all frameworks. However, as the dimensionality of the data increases and the depth increases in real-world applications, the discrepancy of different models will be more significant. For the basic LSTM, corresponding to a depth of \(\delta=1\), the RAM usage is 567MB. In the \ac{ATF}, RAM usage increases to 959MB and 1159MB for depths \(\delta=2\) and \(\delta=3\) respectively, due to the need to store intermediate prediction results for each step. Similarly, the \ac{PF} framework requires 873MB and 1065MB of RAM at \(\delta=2\) and \(\delta=3\) respectively, reflecting a similar pattern of escalating memory demands as depth increases.

Meanwhile, the \ac{SPF}, leveraging its unique supplementary dataset design, consistently maintains RAM usage at 689MB for both \(\delta=2\) and \(\delta=3\). Both \ac{ATF} and \ac{SPF} initially store all data on the CPU and transfer it to the GPU only for training. This strategy reduces continuous memory usage on the GPU. SPF, however, manages lower RAM usage by utilizing a periodically updated supplementary dataset. This method allows SPF to use a flexible data extraction approach for training, where some data are sourced from the original dataset and some from the supplementary dataset, effectively managing memory demand and optimizing data utilization. This strategy ensures efficient long-term prediction capability by integrating predictions back into training without the need for retaining extensive intermediate data in GPU memory, contrary to \ac{ATF} where each step’s result must be preserved.
Although the proof-of-concept in this study is demonstrated on a relatively low-dimensional dynamical system, the \ac{SPF} approach holds significant potential to substantially reduce the computational resources required for real-world high-dimensional dynamical systems, thereby enhancing the frugality of large AI models.

\begin{table}[h]
\centering
\caption{RAM Usage by Different Training Frameworks}
\begin{tabular}{|c|c|c|}
\hline
\textbf{Training Framework} & \textbf{Depth} (\(\delta\)) & \textbf{RAM Usage (MB)} \\
\hline
Basic LSTM & 1 & 567 \\
ATF & 2 & 959 \\
ATF & 3 & 1159 \\
PF & 2 & 873\\
PF & 3 & 1065\\
SPF & 2 & 689 \\
SPF & 3 & 689 \\
\hline
\end{tabular}
\label{tab:ram_usage}
\end{table}

\section{Conclusion}
\label{sec:conclusion}
Achieving long-term predictions through neural networks is a long-standing challenge, especially in the fluid dynamics community. Autoregressive-like training frameworks are often investigated due to their straightforward implementation and their independence from extensive prior knowledge. In this paper, we introduce and implement a novel training framework, \ac{SPF}, inspired by \ac{PF} and designed to address the high GPU memory demands and the lack of frugality typically associated with autoregressive approaches.

\ac{SPF} retains the simplicity of one-step-ahead training while significantly enhancing long-term prediction performance and maintaining computational efficiency. Notably, \ac{SPF} reduces GPU memory usage by up to 40\% compared to autoregressive methods when $\delta = 3$ and ensures scalability, as its memory footprint remains constant regardless of $\delta$. Furthermore, \ac{SPF} achieves longer, reliable predictions, consistently maintaining high performance even over unseen time steps. It also preserves short-term accuracy, ensuring precise immediate predictions while refining its understanding of long-term dynamics through iterative updates. Its robustness is further evidenced by its reliable performance in noisy environments, where \ac{SPF} continues to deliver dependable predictions despite suboptimal input conditions. All the results demonstrating these capabilities are obtained from benchmark cases Burgers' equation and shallow water system, widely used test cases that effectively showcase \ac{SPF}’s enhanced efficiency, stability, and accuracy in complex, time-dependent scenarios. \hao{The influence of hyperparameters on the performance of \ac{SPF} has also been investigated, providing insights into their impact on prediction quality and offering practical guidelines for selecting appropriate values.}

The \ac{SPF}, with its autoregressive architecture, holds substantial promise for both large-scale simulations and complex, real-world scenarios where traditional physical formulas are either inadequate or unavailable. \hao{Under scenarios where prior physical knowledge is available, \ac{SPF} has been shown to efficiently integrate such information to enhance prediction performance.} Nevertheless, \ac{SPF} has limitations. Its use of a supplementary dataset and acquisition method adds complexity compared to standard one-step or autoregressive approaches, requiring additional effort to maintain and update datasets and select training samples. To save GPU memory, datasets must reside on the CPU, and transferring data to the GPU during training can introduce communication overhead and latency. To address this issue, future work could focus on optimizing data transfer and acquisition methods, possibly through asynchronous loading or more efficient compression, to reduce communication overhead. Furthermore, extending and validating \ac{SPF} in various scenarios may help evaluate its robustness and scalability, directing its development towards more effective and broadly applicable solutions.

\section*{Data and code availability}
The code and example structure of models of the \ac{SPF} used in this paper is available at \url{https://github.com/HaoZhou-713/stochastic-pushforward.git}.

\section*{Acknowledgement}

Sibo Cheng acknowledges the support of the French Agence Nationale de la Recherche (ANR) under reference ANR-22-CPJ2-0143-01. Hao Zhou gratefully acknowledges the QUT for support through a Ph.D. scholarship.

\section*{Abbreviations}

\begin{acronym}[AAAAA]
\footnotesize{
\acro{AE}{Autoencoder} 
\acro{ATF}{Autoregressive training framework}
\acro{CAE}{Convolutional autoencoder} 
\acro{DL}{Deep learning}
\acro{FDM}{Finite difference method}
\acro{FVM}{Finite volume method}
\acro{LBM}{Lattice boltzmann method}
\acro{LSTM}{Long short-term memory}
\acro{ML}{Machine learning}
\acro{MSE}{Mean square error}
\acro{PDE}{Partial differential equation}
\acro{PF}{PushForward}
\acro{RNN}{Recurrent neural network}
\acro{ROM}{Reduced-order modelling}
\acro{SSIM}{Structural similarity index}
\acro{Seq2Seq}{Sequence-to-Sequence}
\acro{SPF}{Stochastic PushForward}
}
\end{acronym}

\newpage
\footnotesize
\bibliographystyle{elsarticle-num-names}
\bibliography{references}  

\end{document}